\documentclass[lettersize,journal]{IEEEtran}
\usepackage{amsmath,amsfonts}
\usepackage{algorithmic}
\usepackage{array}
\usepackage[caption=false,font=footnotesize,labelfont=rm,textfont=rm]{subfig}
\usepackage{textcomp}
\usepackage{stfloats}
\usepackage{url}
\usepackage{verbatim}
\usepackage{graphicx}
\usepackage{booktabs}
\usepackage{subfig} 
\usepackage{multirow} 
\usepackage{doi}

\def\BibTeX{{\rm B\kern-.05em{\sc i\kern-.025em b}\kern-.08em
    T\kern-.1667em\lower.7ex\hbox{E}\kern-.125emX}}
\usepackage{balance}
\begin{document}
\title{A Text-Based Knowledge-Embedded Soft Sensing Modeling Approach for General Industrial Process Tasks Based on Large Language Model}
\author{Shuo~Tong, Han~Liu,~\IEEEmembership{Member,~IEEE,} Runyuan~Guo,~\IEEEmembership{Member,~IEEE,} Xueqiong~Tian, Wenqing~Wang,~\IEEEmembership{Member,~IEEE,} Ding~Liu,~\IEEEmembership{Member,~IEEE,} and~Youmin~Zhang,~\IEEEmembership{Fellow,~IEEE}
\thanks{This work was supported in part by the National Natural Science Foundation of China under Grant 92270117 and Grant 62376214, in part by the Natural Science Basic Research Program of Shaanxi under Grant 2024JC-YBQN-0697 and Grant 2023-JC-YB-533. \textit{(Corresponding author: Han Liu.)}}
\thanks{Shuo Tong, Han Liu, Runyuan Guo, Xueqiong Tian, Wenqing Wang, and Ding Liu are with the Xi’an University of Technology, Xi’an 710048, China (e-mail: ts842029379@gmail.com; liuhan@xaut.edu.cn; guorunyuan@xaut.edu.cn; tianxueqiong@stu.xaut.edu.cn; wangwenqing@xaut.edu.cn; liud@xaut.edu.cn).}
\thanks{Youmin Zhang is with the Department of Mechanical, Industrial, and Aerospace Engineering and the Concordia Institute of Aerospace Design and Innovation, Concordia University, Montreal, QC H3G 1M8, Canada (e-mail: ymzhang@encs.concordia.ca).}
\thanks{Color versions of one or more of the figures in this article are available online at http://ieeexplore.ieee.org.}

}

\markboth{IEEE TRANSACTIONS ON XXX}%
{How to Use the IEEEtran \LaTeX \ Templates}

\maketitle

\begin{abstract}
Data-driven soft sensors (DDSS) have become mainstream methods for predicting key performance indicators in process industries. However, DDSS development requires complex and costly customized  designs tailored to various tasks during the modeling process. Moreover, DDSS are constrained to a single structured data modality, limiting their ability to incorporate additional contextual knowledge. Furthermore, existing DDSSs' limited representation learning leads to weak predictive performance with scarce data. To address these challenges, we propose a general framework named LLM-TKESS (large language model for text-based knowledge-embedded soft sensing), harnessing the powerful general problem-solving capabilities, cross-modal knowledge transfer abilities, and few-shot learning capabilities of LLM for enhanced soft sensing modeling. Specifically, an auxiliary variable series encoder (AVS Encoder) is proposed to unleash LLM's potential for capturing temporal relationships within series of process data and spatial semantic relationships among auxiliary variables. Then, we propose a two-stage fine-tuning alignment strategy: in the first stage, employing parameter-efficient fine-tuning (PEFT) through autoregressive training adjusts LLM to rapidly accommodate process variable data, resulting in a soft sensing foundation model (SSFM). Subsequently, by training adapters, we adapt the SSFM to various downstream soft sensing tasks without modifying its architecture. Then, we propose two text-based knowledge-embedded soft sensors, integrating new natural language modalities to overcome the limitations of pure structured data models. Furthermore, benefiting from LLM's pre-existing world knowledge, our model demonstrates outstanding predictive capabilities in small sample conditions. Finally, using the thermal deformation of air preheater rotor as a case study, we validate through extensive experiments that LLM-TKESS exhibits outstanding performance.
\end{abstract}


\section{Introduction}
\IEEEPARstart{I}{n} modern industrial processes, accurate and stable measurement of key quality variables plays a critical role in real-time process monitoring and optimization, product quality control, and energy conservation. However, with the continuous expansion of industrial scale, processes are becoming increasingly complex and the number of process variables is growing. Some critical variables cannot be effectively monitored in real-time using physical hardware due to uncertainties in measurement environments, high maintenance costs of detection instruments, and high latency in offline measurements \cite{9716776, HE2023105737}. To address this issue, soft sensing technology has emerged. This technology establishes predictive models based on easily measurable process variables (auxiliary variables) and mathematical relationships with difficult-to-measure quality variables (primary variables), providing rapid quality variable information for production processes. Due to the advantages of low cost and high performance, soft sensors is widely studied and applied in various industrial fields \cite{9794453}.

Currently, there are two main modeling approaches for soft sensing: mechanism-driven models and data-driven models \cite{GE201716,10329566}. Due to the complexity of industrial processes, it is often challenging to obtain sufficient prior mechanistic knowledge for accurate mechanism-driven modeling \cite{9447941}. In contrast, data-driven models do not require prior expert knowledge and can construct accurate prediction models solely based on extensive historical data. In recent years, fueled by the robust feature representation and nonlinear fitting capabilities of deep learning, data-driven soft sensor (DDSS) based on deep learning have emerged as the mainstream approach in soft sensing modeling, achieving state-of-the-art (SOTA) performance \cite{9329169}. The popular paradigms in DDSS model include long short-term memory (LSTM) \cite{HE2023105737}, autoencoder and its variants \cite{9716776,10329566}, and convolutional neural network (CNN) \cite{9447941}. 
\begin{figure*}[!t]
\centering
\includegraphics[width=5.2in]{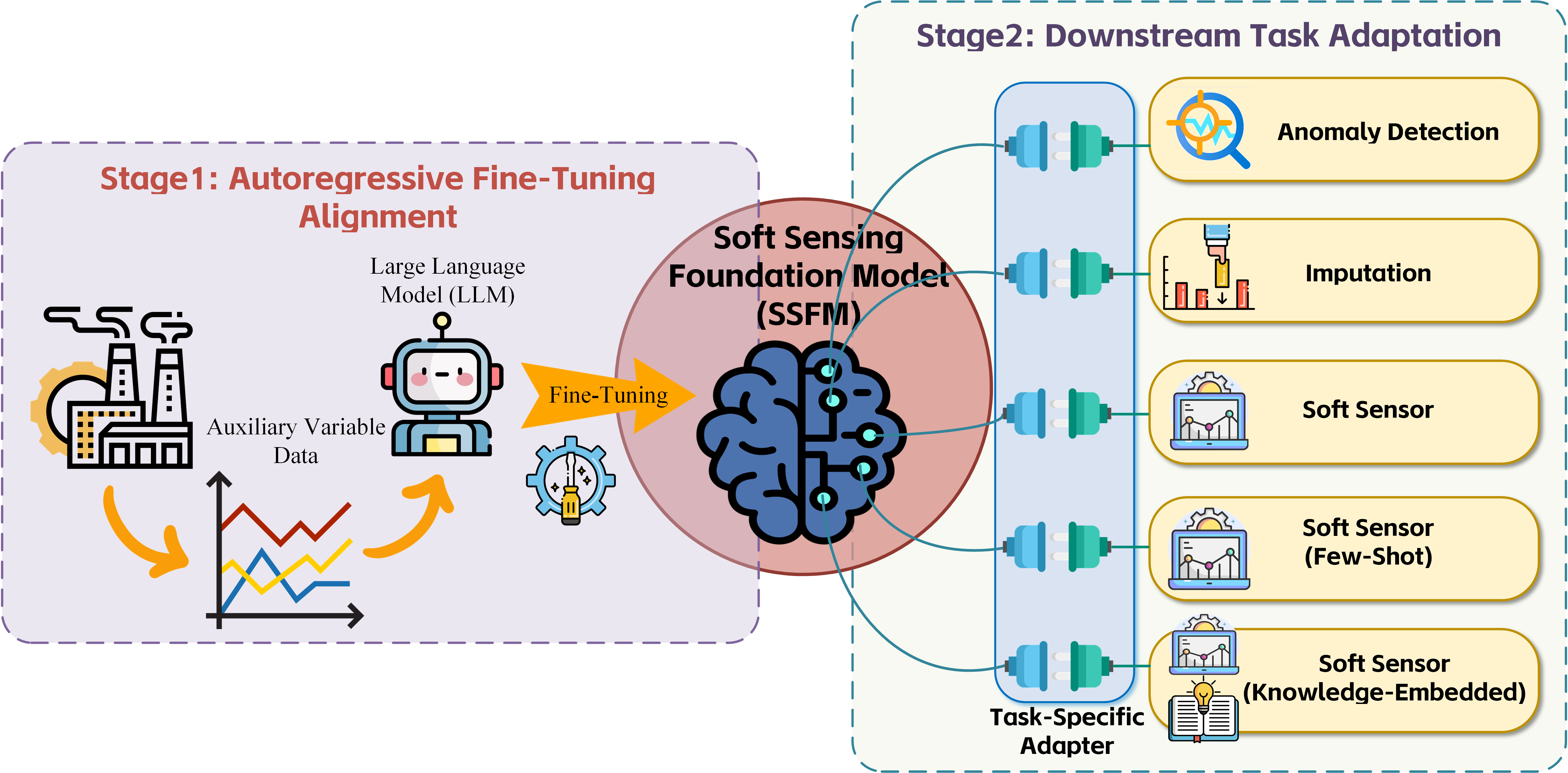}
\caption{Conceptual framework of soft sensing foundation model (SSFM) based on LLM.}
\label{fig1}
\end{figure*}

Despite the significant achievements of deep learning-based soft sensing methods to date, the current approaches still face three major challenges that have yet to be overcome: {\bf{\textit{Limited universality.}}} The soft sensing pipeline is a serial, multi-stage, multi-task modeling process \cite{GE201716,Vallejo2019,SOUZA201669,WANG2018112}. Apart from final soft sensor modeling stages, tasks like anomaly detection and missing value imputation are crucial for model quality \cite{9235582,ZHU2018107}. However, the fragmented nature of these multi-stage tasks requires data analysts to employ customized deep learning methods for each stage, involving separate modeling, parameter tuning, and optimization, which significantly increases the development cost and complexity of the models. Thus, current soft sensing modeling typically relies on simple statistical techniques for data preprocessing or neglects the anomaly handling steps \cite{9794453,WANG2018112}. As discussed in Section \ref{Data Preprocessing in Soft Sensing Tasks}, such inaccurate modeling approaches pose significant risks to the reliability and safety of soft sensors. {\bf{\textit{Limited input modality.}}} DDSS are restricted to using only structured data from industrial processes as input \cite{9716776,HE2023105737,9794453}. The restricted contextual information prevents the models from conducting a comprehensive analysis from multiple perspectives and levels, hindering the enhancement of robustness and representational capacity. Moreover, structured data lacks semantic-level descriptions and rich contextual or background information, resulting in data-driven models' inability to interpret high-level abstract concepts \cite{9794453}. {\bf{\textit{Poor few-shot learning capability.}}} Due to the uncertainty and the high measurement costs associated with certain key variables, the number of samples available for engineering applications is often limited \cite{9454563,10233112}. This presents a challenge for traditional DDSS. Existing soft sensing approaches for few-shot tasks are typically categorized into three main types: gray-based methods \cite{WANG2014383}, feature extraction-based methods \cite{dernoncourt2014analysis}, and the virtual sample generation-based methods \cite{10233112}. However, all three approaches require complex feature engineering or model design specifically, which is time-consuming and requires domain expertise.

Traditional DDSS struggle to effectively address these challenges. Recently, large-scale pre-trained language models (LLMs), owing to their vast number of parameters and extensive training data, have demonstrated exceptional performance across a wide range of downstream tasks in both natural language processing (NLP) and computer vision (CV) \cite{zhao2023survey}. This success is largely attributed to the emergent capabilities (unexpected new abilities) of LLM, such as robust generalization, cross-modal knowledge transfer, multimodal capabilities, and knowledge representation \cite{wei2022emergent,10386743}. These features make the development of multimodal artificial general intelligence (AGI) a viable possibility. This paper aims to introduce LLM to explore whether the emergent capabilities can address the challenges faced by DDSS. The specific details are as follows:

{\bf{Universality}}: Pre-trained LLMs are currently among the most popular used foundation models (also known as large models). Foundation models (FMs) are powerful general-purpose models trained on vast amounts of data to generate general-purpose representations \cite{zhou2023comprehensive}. These representations are subsequently fine-tuned to facilitate model training across various downstream tasks. Compared to traditional training paradigms that require the specialized design of algorithms for each task, this new universal paradigm reduces the cost and complexity associated with developing task-specific models. As a result, universal FMs have emerged across multiple domains such as NLP \cite{myers2024foundation}, CV \cite{shah2023vint}, time-series forecasting \cite{zhou2023fits}, and medicine \cite{moor2023foundation}. 

As illustrated in Fig. \ref{fig1}, we establish a universal soft sensing foundation model (SSFM) adapted for structured data through fine-tuning a pre-trained LLM. This highly compatible SSFM is coupled with various task-specific adapters (TSAs) in downstream tasks to provide robust computational, learning, and problem-solving capabilities at the foundational level, enabling rapid data processing initiation. This transformation facilitates the shift from artisanal soft sensing models to a factory-style foundational model paradigm, meeting diverse operational needs in industrial processes efficiently.

{\bf{Cross-modal knowledge transfer and multimodal}}: The underlying architecture of FMs, transformer, can serialize modalities such as text, images, videos, and audio into a unified set of tokens for training, enabling multimodal universality and cross-modal knowledge transfer \cite{NIPS2017_3f5ee243}. For example, Voice2Series \cite{DBLP:journals/corr/abs-2106-09296} aligns time series data with acoustic model (AM) formats and leverages large-scale pre-trained AM for time series classification tasks. Bao et al. introduced a universal vision-language pretrained Model (VLMo) \cite{DBLP:journals/corr/abs-2111-02358}, fine-tuning visual models to train language experts. ST-LLM \cite{https://doi.org/10.1155/2019/8392592} utilizes LLM with a novel attention freeze strategy to achieve optimal performance in traffic prediction. 

The emergence of such cross-modal capabilities lies in the essential ability of transformers to capture dependencies between different tokens across long sequences, as well as the weight-sharing mechanism \cite{DBLP:journals/corr/abs-2106-09296,10386743}. Therefore, the presence of contextual correlations between sequences is key to determining whether this modality can be successfully handled using LLMs. Given that process variable data exhibits strong temporal relationships, which align with this criterion, we have specifically designed an auxiliary variable series encoder (AVS Encoder) for industrial process auxiliary variables (AVs). AVS Encoder tokenizes and embeds AVs data to adapt it to the input requirements of LLM, leveraging LLM's capability in representation learning to rapidly adapt to process variable data. In addition, we aim to introduce new input modalities for DDSS through the cross-modal capabilities of the SSFM, enriching the input representations. To this end, we have designed two text-based knowledge-embedded soft sensors (TKESS), namely the prompt-driven soft sensor (LLM-PSS) and the prompt and data mixed embedding-driven soft sensor (LLM-PDSS). Introducing knowledge into soft sensors in the form of natural language has three advantages: (1) Enriching the input representations of soft sensor, the introduction of knowledge through prompt-driven learning further enhances the comprehensiveness and performance of the model, and provides new research directions in the field of soft sensing. (2) Compared to data-driven models, natural language is more in line with human habits and preferences, making LLM-PSS more user-friendly and easy to access for non-researcher users. (3) Compared to data-driven models, explanations in natural language are more in line with human cognitive habits and logical reasoning, enhancing the intuitiveness and information content of model interpretations.

{\bf{Powerful knowledge representation capability}}: LLMs are trained through self-supervised learning on vast amounts of data without external labels. This enables LLMs to extract richer and higher-level knowledge representations from the data \cite{zhao2023survey}. Harnessing these learned knowledge representations, LLMs demonstrate powerful reasoning and pattern recognition abilities, primarily manifested in two aspects: (1) Knowledge transfer capability, where LLMs effectively transfer pre-trained knowledge to downstream tasks within the same modality or across modalities \cite{DBLP:journals/corr/abs-2106-09296,DBLP:journals/corr/abs-2111-02358,https://doi.org/10.1155/2019/8392592}. (2) Few-shot learning capability, where LLMs have been proven to excel even in scenarios with limited or zero labeled examples \cite{NEURIPS2020_1457c0d6}. In this paper, we leverage these two knowledge representation capabilities of LLMs to further enhance the limited predictive accuracy of traditional DDSS and to tackle the challenge of insufficient in-domain data for small sample scenarios.

In summary, based on pre-trained LLM, we introduce for the first time a universal two-stage adaptation fine-tuning strategy termed LLM-TKESS (LLM for text-based knowledge-embedded soft sensing), for general soft sensing modeling tasks, aiming to unleash the potential of LLM in addressing key challenges in traditional soft sensing tasks. The main contributions of this paper can be summarized as follows:
\begin{enumerate}{}{}
\item{The AVS Encoder was proposed by us, designed with a novel tokenization and embedding method based on the horizontal temporal and vertical spatial characteristics of AVs. This aims to bridge the modality gap between natural language modalities and process variable data.}
\item{We introduce a two-stage universal soft sensing modeling framework, LLM-TKESS. Without distorting the inherent representations of LLM, we first conduct autoregressive PEFT to train the SSFM aligned with AVs tokens. Subsequently, we design TSAs tailored for different downstream tasks and seamlessly integrate them with SSFM in a lightweight manner.}
\item{In the downstream tasks, two different TKESS, LLM-PSS and LLM-PDSS, were introduced by us based on prompt learning, representing our initial attempt to incorporate natural language modalities into soft sensor.}
\item{Through extensive experiments, we have demonstrated that the powerful representation learning capabilities of LLMs effectively enhance the prediction accuracy of soft sensor, achieving SOTA performance. Additionally, we validate the robust few-shot learning capabilities by leveraging the innate knowledge of LLMs without the need for additional complex operations.}
\end{enumerate}

\section{Preliminaries}
\subsection{Transformer in LLMs}
LLMs exhibit various evolutions and types, but fundamentally, they employ the transformer architecture as their underlying framework. One of the core components of the transformer is the multi-head self-attention (MSA) mechanism. MSA simultaneously computes multiple self-attention mechanisms, capturing rich long-range dependencies within sequences across various subspaces. This endows LLMs with significant contextual comprehension, enabling them to adeptly manage diverse multimodal tasks. The specific pipeline of the MAS is shown in the gray area in Fig. \ref{fig3}(a). Suppose the $t\text{-th}$ input sequence is denoted by ${{X}_{t}}\in {{\mathbb{R}}^{n\times d}}$, where $n$ represents the sequence length and $d$ symbolizes the embedding dimension. Assumed a total of $h$ attention heads, for every individual attention head $j$, the input transformation into ${{Q}_{j}}\in {{\mathbb{R}}^{n\times {{d}_{k}}}}$ (query), ${{K}_{j}}\in {{\mathbb{R}}^{n\times {{d}_{k}}}}$ (key), and ${{V}_{j}}\in {{\mathbb{R}}^{n\times {{d}_{k}}}}$ (value) via the respective linear mapping matrices $W_{q}^{j}\in {{\mathbb{R}}^{d\times {{d}_{k}}}}$, $W_{k}^{j}\in {{\mathbb{R}}^{d\times {{d}_{k}}}}$ and $W_{v}^{j}\in {{\mathbb{R}}^{d\times {{d}_{k}}}}$.
\begin{equation}
{{Q}_{j}}={{X}_{t}}W_{q}^{j},\text{ }{{K}_{j}}={{X}_{t}}W_{k}^{j},\text{ }{{V}_{t}}={{X}_{t}}W_{v}^{j}
\end{equation}

The dot product of the query ${{Q}_{j}}$ and key ${{K}_{j}}$ is transformed into attention weights via the softmax function and subsequently, these weights are used in a weighted summation with the value ${{V}_{j}}$ to calculate the output $hea{{d}_{j}}$ of self-attention for each head.
\begin{small} 
\begin{equation}
\text{ }\!\!~\!\!\text{ hea}{{\text{d}}_{j}}\text{=Attention}\left( {{Q}_{j}},{{K}_{j}},{{V}_{j}} \right)=\text{Softmax}\left( \frac{{{Q}_{j}}K_{j}^{T}}{\sqrt{{{\text{d}}^{\text{k}}}}} \right){{V}_{j}}
\end{equation}
\end{small}
Subsequently, the outputs of self-attention from all heads are aggregated through a concatenation operation. The aggregated result then undergoes a linear transformation via matrix ${{W}_{o}}\in {{\mathbb{R}}^{d\times d}}$, culminating in the final output of MSA:
\begin{equation}
\resizebox{0.9\hsize}{!}{$\begin{aligned}
\text{Multi-head}\left( Q,K,V \right)=\text{Concat}\left( \text{hea}{{\text{d}}_{1}},\text{hea}{{\text{d}}_{2}},...,\text{hea}{{\text{d}}_{h}} \right){{W}_{o}}
\end{aligned}$}
\end{equation}

LLMs construct the network architecture through the stacking of multiple layers of transformers. 

\subsection{Pre-Training LLMs with Autoregressive Training Method}

An additional significant reason for the potent representation learning capability of pre-trained LLMs originates from the utilization of expansive datasets to implement autoregressive training approaches for the sequential synthesis of text. In autoregressive training, assuming a given sequence $X\in {{\mathbb{R}}^{n\times d}}$ contains n tokens ${{x}_{1}},{{x}_{2}},...,{{x}_{n}}$, the LLM is required to predict the next token ${{x}_{n+1}}$ in the sequence based on the given input. Upon obtaining this element, it is appended to the input sequence to form a new model input for continued iterative prediction. LLMs typically employ the log-likelihood as the loss function for autoregressive training. The objective function of LLM's pre-training can be formulated as follows:
\begin{equation}
{{\mathcal{L}}_{PLM}}\left( X \right)=-\sum\limits_{i=1}^{n}{\log P\left( {{x}_{i}}|{{x}_{1}},{{x}_{2}}...,{{x}_{i-1}} \right)}
\end{equation}

This self-supervised learning approach empowers LLMs to unearth richer and more advanced-level features from the data during feature representation learning. Consequently, LLMs acquire powerful cross-modal knowledge transfer capabilities, which create the necessary conditions for LLMs to effectively tackle diverse data tasks.

\begin{figure}[!t]
\centering

\includegraphics[width=3in]{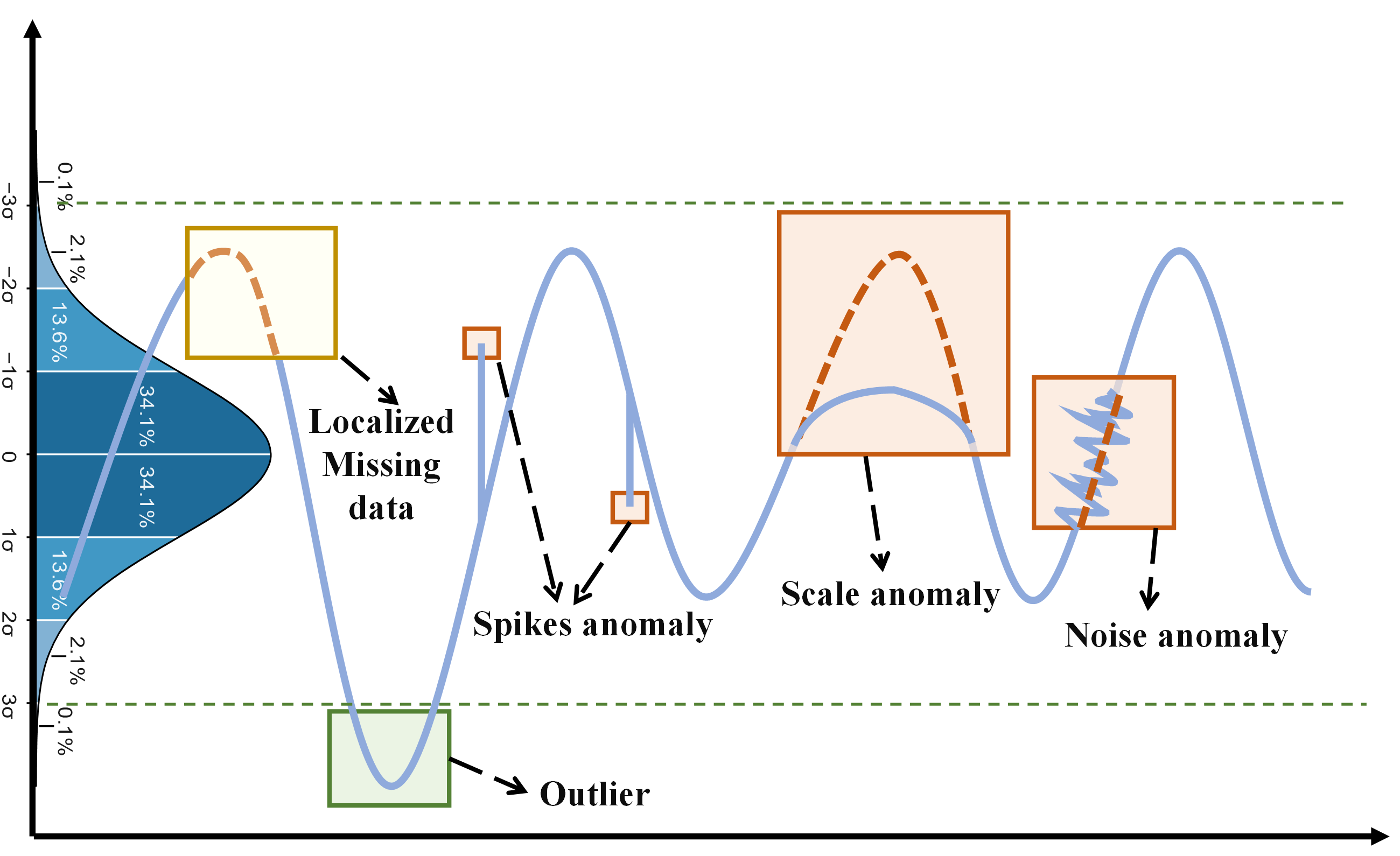}

\caption{Different types of anomalies and missing data.}
\label{fig2}
\end{figure}

\subsection{Data Preprocessing in Soft Sensing Tasks}\label{Data Preprocessing in Soft Sensing Tasks}

Possessing comprehensive and accurate data forms a crucial precondition for the accurate modeling of DDSS \cite{9584888}. However, due to system failures, sensor malfunctions, or other factors, the collected industrial data often contains a significant amount of anomalies and missing values, which severely impair the generalization ability of DDSS, greatly compromising their safety and hindering their deployment and application \cite{9454563,WANG2014383,dernoncourt2014analysis}. Therefore, it necessitates relevant preprocessing operations on the raw data, specifically for detecting anomalies and imputing missing values to enhance data quality and consistency.

Currently, to streamline the modeling process, most soft sensing methods conventionally opt for statistical models to execute rudimentary preprocessing of collected data. For instance, as depicted in Fig. \ref{fig2}, numerous methods remove outliers using box plots or the $\text{3}\sigma $ rule \cite{9794453,WANG2018112}. Nevertheless, statistical techniques fail to account for temporal information and contextual relations within the data, merely capable of detecting outliers that fall outside the range of data distribution (green box). They are ineffectual at accurately detecting anomalies such as spikes anomaly from instantaneous electromagnetic interference from industrial sensors, noise anomaly ensuing from environmental noise, and scale anomaly due to calibration errors (red box). However, these three types of anomalies are frequently encountered in industrial data collection. In addition, while traditional imputation methods might fulfill single missing data by calculating the mean \cite{Vallejo2019}, statistical methods struggle to accurately align and impute data when confronted with localized missing sections (yellow box). Based on deep learning models, the detection of anomalies and data imputation in industrial data can achieve better performance \cite{WANG2014383,dernoncourt2014analysis,MA2024121428}, but this requires the design and separate training of different complex deep learning models specifically for data preprocessing and soft sensing in various industrial data scenarios, which significantly increases the workload and the difficulty of modeling.

Hence, establishing a robust and accurate general soft sensing framework applicable across stages is crucial for the precise and efficient deployment of soft sensors. Given LLM's robust versatility and generalization capabilities across multiple tasks, we utilize fine-tuned LLM as the SSFM. By incorporating minimal-parameter adapters for anomaly or missing value handling, we achieve precise treatment of various types of anomalies and gaps without the need for additional deep learning modeling.

\begin{figure*}[t]
\centering
\includegraphics[width=5.5in]{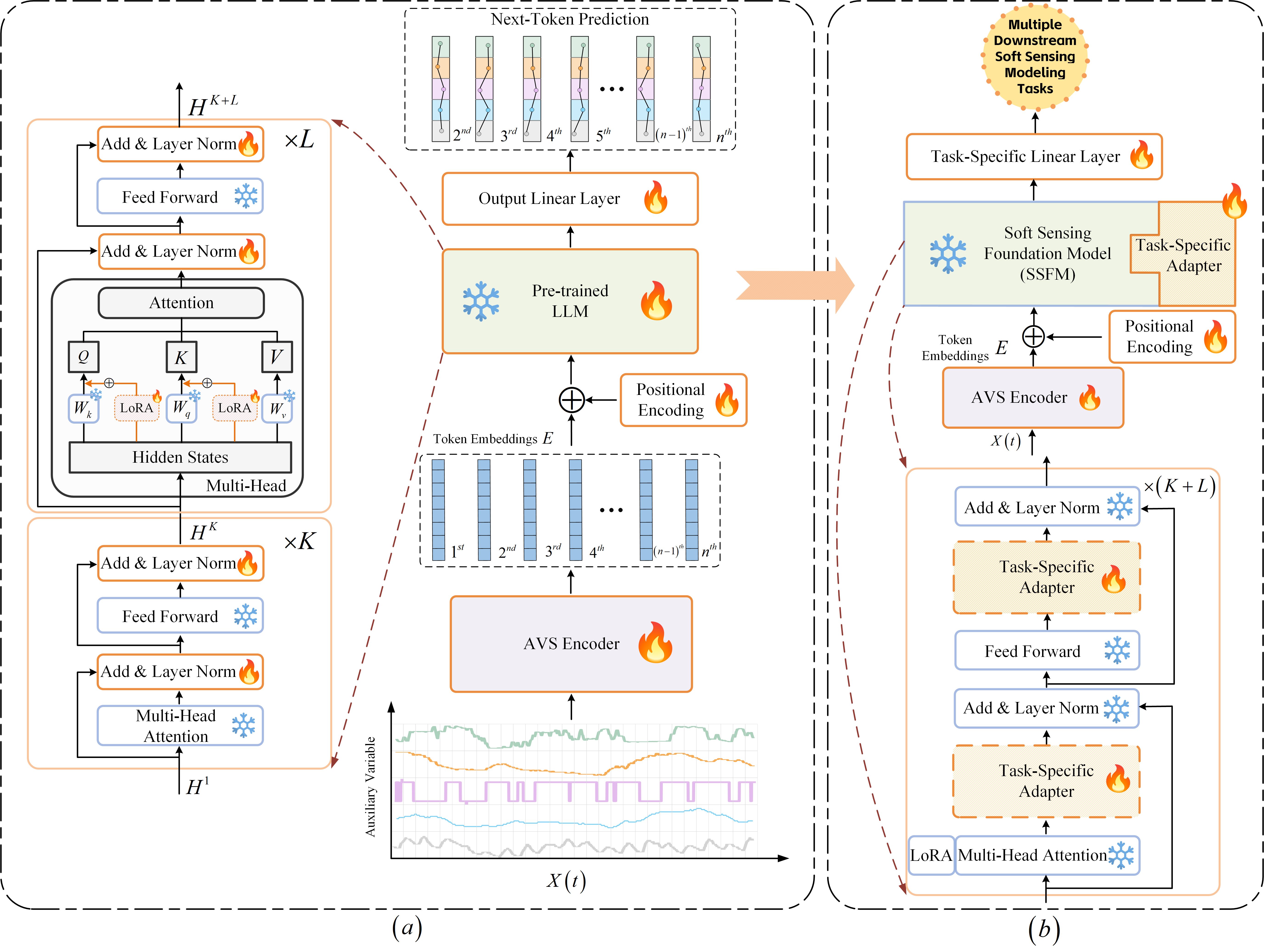}
\caption{The proposed the overall framework architectures of LLM-TKESS. (a) Structure of SSFM autoregressive fine-tuning alignment phase. (b) Structure of SSFM-TSA downstream task adaptation phase.}
\label{fig3}
\end{figure*}

\subsection{Data Preparation for Time-Lags}\label{3.4}

Constructing inputs that carry sequential information is one of the conditions that catalyze the emergent capabilities of LLMs. In complex industrial processes, the variable data measured by sensors exhibit strong dynamic correlations over time \cite{9794453}. Constructing multivariable input data with time lags allows LLM to consider not only the vertical correlations among AVs but also capture the horizontal semantic dependencies over time. This maximizes the utilization of the inherent semantic information within the data and significantly leverages the long-distance knowledge representation inherent in pre-trained LLM. For the time-lagged input data at time $t$, $X\left( t \right)\in {{\mathbb{R}}^{n\times m}}$ can be represented as:
\begin{equation}
\begin{aligned}
  & X(t)=\left[ x(t),\text{ }x(t-1),\text{ }x(t-2),\text{ }\ldots ,\text{ }x\left( t-n+1 \right) \right] \\ 
 & =\left[ \begin{matrix}
   {{x}_{1}}(t), & {{x}_{1}}(t-1), & \cdots , & {{x}_{1}}(t-n+1)  \\
   {{x}_{2}}(t), & {{x}_{2}}(t-1), & \cdots , & {{x}_{2}}(t-n+1)  \\
   \vdots  & \vdots  & \vdots  & \vdots   \\
   {{x}_{m}}(t), & {{x}_{m}}(t-1), & \cdots , & {{x}_{m}}(t-n+1)  \\
\end{matrix} \right] \\ 
\end{aligned}
\end{equation}
where $n$ denotes the size of the historical time series window, and $m$ represents the dimension of the AVs. Multivariate sequence samples $X\left( t \right)$ are used as model inputs during both stages of LLM-TKESS.

\section{Methodology}
\subsection{Overview Framework}

As depicted in Fig. \ref{fig3}, the proposed LLM-TKESS consists of two primary stages: the self-supervised fine-tuning alignment stage and the SSFM-TSA downstream task adaptation stage. In the first stage, the proposed AVS Encoder is used to converts the time series of normalized multivariate into token embeddings feature representations suitable for pre-trained LLM input. These token embeddings are then fed into the pre-trained LLM along with a learnable positional encoding. Through autoregressive training, the pre-trained LLM is fine-tuned to align industrial sensor data with the LLM's knowledge representation, resulting in the soft sensing foundation model (SSFM). In the second stage, SSFM serves as the base model for various downstream tasks. Depending on the specific soft sensing tasks at hand, a task-specific adapter (TSA) is selected and integrated with SSFM. While maintaining all SSFM parameters frozen, various downstream tasks are trained by fine-tuning only a small set of adapter parameters. Finally, capitalizing on the LLM's multimodal capabilities, two novel TKESS are introduced, expanding the possibilities within the soft sensing modeling domain.

\subsection{Proposed AVS Encoder}\label{4.2}
The key to enabling LLMs achieve cross-modal knowledge transfer lies in constructing a downstream task token embedding that is compatible with the LLM. During the pretraining process, LLM's tokenizer divides natural language sequences into discrete tokens at the word, subword, or character level. However, in industrial modeling, the inputs of soft sensor time-lags auxiliary variable (AVs)—and the outputs primary variable are time series. Traditional discrete tokenizers struggle to represent time series accurately and finely \cite{gruver2023large}. Based on this, considering the multivariate dependency characteristics of AVs, we propose a new auxiliary variable series encoder (AVS Encoder) to perform tokenizer and high-dimensional embedding on the multivariate series samples obtained in Section \ref{3.4}, to better adapt to the knowledge representation of LLMs.

\begin{figure*}[!t]
\centering
\includegraphics[width=5.5in]{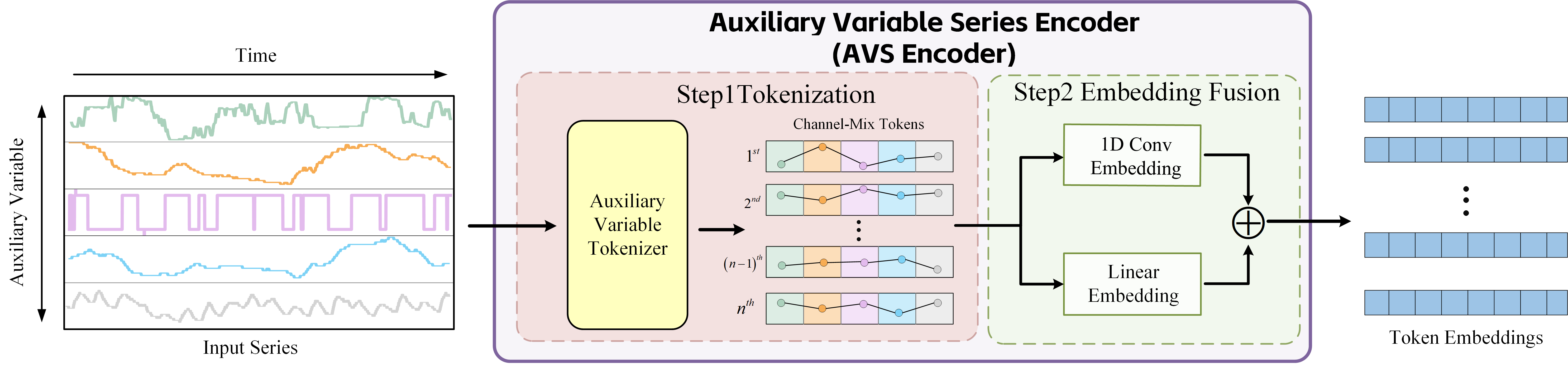}
\caption{Structure of the proposed AVS Encoder.}
\label{fig4}
\end{figure*}

Fig. \ref{fig4} illustrates the specific structure of the AVS Encoder. Firstly, to address the strong coupling and non-linear relationships among complex AVs, we design an auxiliary variable tokenizer suitable for times series data to achieve the tokenization of input samples. Specifically, dividing each individual time step as a unit, the input sample $X\left( t \right)$ is divided into $n$ tokens, denoted as ${{T}_{1}},{{T}_{2}}\cdots {{T}_{n}}$, to better capture the longitudinal non-linear relationships among AVs. Each token contains all auxiliary variable values at that moment. The $p$ token ${{T}_{p}}\in {{\mathbb{R}}^{m}}$ (where$p=1\ldots n$ ) can be represented as follows:
\begin{equation}
{{T}_{p}}=\left[ {{x}_{1}}\left( p \right),\text{ }{{x}_{2}}\left( p \right),\text{ }\ldots {{x}_{m}}\left( p \right) \right]
\end{equation}

This tokenization approach yields a number of tokens equivalent to the time window, with the tokens also exhibiting horizontal temporal relationships. This approach better captures the complex cross-channel information interaction and related patterns among AVs, thereby further enhancing the model's performance. Subsequently, through the operation of embedding fusion, all tokens are embedded into a continuous high-dimensional vector space to adapt to the inputs of LLMs. To more effectively capture the semantic information of input tokens, we employ two branches: one utilizing 1D convolution to extract local feature dependencies, and the other using linear probing to capture global deep features. Finally, the extracted features from these two branches are fused to obtain the token embeddings $E\in {{\mathbb{R}}^{n\times d}}$, where $d$ represents the embedding dimension.

\subsection{SSFM Autoregressive Fine-Tuning Alignment Phase}\label{3.3}

During the pre-training phase of LLMs, self-supervision is typically employed to train these models on a substantial corpus of unlabeled textual data. However, soft sensing tasks in industrial settings involve non-language data, which limits the adaptability of LLMs to downstream industrial tasks. Therefore, it is crucial to bridge the gap between the upstream and downstream domains using a data-efficient approach. One of the most common self-supervised pre-training methods for LLMs is autoregressive language modeling, where the objective is to predict the next token in a sequence. Consequently, LLMs exhibit inherent autoregressive properties \cite{han2024parameterefficient}. Research has demonstrated that employing an autoregressive training strategy similar to the one used during the pretraining phase can effectively enhance performance in downstream tasks \cite{10.5555/3600270.3602281,jin2024timellm}. Based on this, we adopted a similar self-supervised autoregressive training approach for the channel-mix tokens obtained in Section \ref{4.2} during the first phase of training the LLM-TKESS. This approach aims to leverage the intrinsic knowledge of LLM to better acquaint them with the characteristics of industrial data. To ensure that the data-independent representation learning capability of the LLM is not compromised during this process, we employed parameter efficient fine-tuning (PEFT) \cite{han2024parameterefficient}. This technique involves freezing the majority of model parameters and selectively fine-tuning only a minimal subset of parameters, thereby aligning the model with downstream tasks and yielding the SSFM.

\begin{figure}[!t]
\centering
\includegraphics[width=3in]{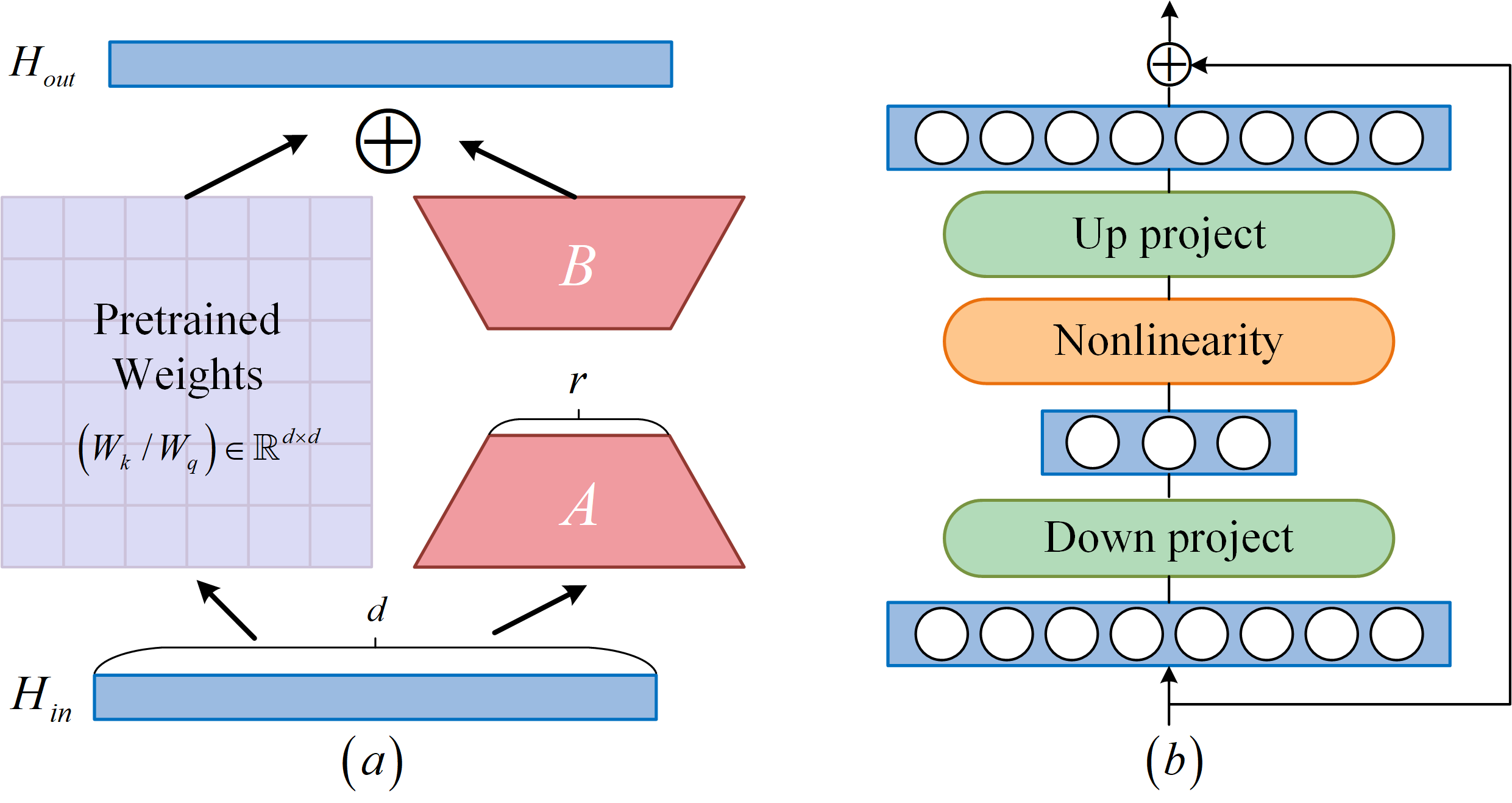}
\caption{Structure of LoRA and task-specific adapter (TSA). (a) Structure of LoRA. (b) Structural of TSA.}
\label{fig5}
\end{figure}

The diagram illustrating the autoregressive fine-tuning alignment stage of the SSFM is depicted in Fig. \ref{fig3}(a). This stage integrates the AVS Encoder, positional encoding, the pre-trained LLM, and an output linear layer. Initially, an input sequence $X\left( t \right)\in {{\mathbb{R}}^{n\times m}}$, comprising $n$ temporal steps and $m$ AVs, is processed through the AVS Encoder to generate $n$ token embeddings $E\in {{\mathbb{R}}^{n\times d}}$. These embeddings are then summed with the positional encoding ${{E}_{p}}\in {{\mathbb{R}}^{n\times d}}$ and input into the pre-trained LLM containing $L+K$ layers of transformers. Within the pre-trained LLM, while ensuring that all other parameters remain frozen, we first fine-tune all layer normalizations across the layers using standard practice \cite{lu2022frozen}. Additionally, for the first $K$ layers, parameters of the multi-head attention mechanism are frozen to preserve the LLM's generalization capabilities. Conversely, for the subsequent $L$ layers, we integrate low-rank adaptation (LoRA) \cite{hu2022lora} to each layer, injecting a trainable rank-factorization matrix with a smaller number of parameters into the query and key of the transformer layer of LLM. This adaptation helps to learn the context information between tokens more effectively without undermining the model's inherent expressive power \cite{hu2022lora}.

As illustrated in Fig. \ref{fig5}(a), for the LLM's pre-trained weight matrices ${{W}_{q}}\in {{\mathbb{R}}^{d\times {{d}_{k}}}}$ or ${{W}_{k}}\in {{\mathbb{R}}^{d\times {{d}_{k}}}}$, LoRA introduces two low-rank matrices $A\in {{\mathbb{R}}^{r\times {{d}_{k}}}}$ and $B\in {{\mathbb{R}}^{d\times r}}$ to perform parallel computations with ${{W}_{q}}$ and ${{W}_{k}}$. Taking the weight matrix ${{W}_{q}}$ as an example, with ${{H}_{in}}\in {{\mathbb{R}}^{n\times d}}$ representing the input, LoRA introduces learned weight correction terms $\Delta W$ during the fine-tuning process to encapsulate task-specific knowledge. This results in the output as follows:
\begin{equation}
{{H}_{LoRA}}={{H}_{in}}{{W}_{q}}+\frac{\alpha }{r}{{H}_{in}}\Delta W={{H}_{in}}{{W}_{q}}+\frac{\alpha }{r}{{H}_{in}}BA
\end{equation}

Where $\alpha$ is the scaling factor, during the parameter initialization phase, $A$ is initialized using a random gaussian distribution while $B$ is initialized to a zero matrix. After passing through $L+K$ layers, the input embeddings yield the LLM's output ${{H}^{L+K}}\in {{\mathbb{R}}^{n\times d}}$. Subsequently, we utilize linear probing to map ${{H}^{L+K}}$ to an autoregressive output predicting the next token value. It is noteworthy that in contrast to employing log-likelihood as the loss function during the pre-training phase, we opt for mean squared error (MSE) as the loss function in the initial fine-tuning stage to ensure a continuous and precise representation of the predictive results. Assuming the output of the linear layer is represented by $SSFM\left( \cdot  \right)$ and the training dataset comprises $z$ input samples, the loss function can be articulated as follows:
\begin{equation}
\resizebox{0.9\hsize}{!}{$\begin{aligned}
{{\mathcal{L}}_{SSFM}}=\frac{1}{n}\frac{1}{z}\sum\limits_{z}^{1}{\sum\limits_{i=1}^{n}{{{\left\| SSFM\left( x_{1}^{z},x_{2}^{z}...,x_{i-1}^{z} \right)-x_{i}^{z} \right\|}^{2}}}}
\end{aligned}$}
\end{equation}

\subsection{Adaptation Stage for Downstream Tasks in SSFM-TSA}

In the second stage, we utilize SSFM as the universal base model for downstream task adaptation, fixing all its model parameters. By simply adding plug-and-play task-specific adapters (TSAs) \cite{DBLP:journals/corr/abs-1902-00751} to SSFM, we can effectively model various tasks in soft sensing. Compared to using fine-tuning methods at this stage, where separate entirely new models are trained for each downstream task, this paradigm of base FMs + adapters training ensures high parameter sharing. It requires attaching very few trainable parameters specifically for each downstream task, significantly reducing modeling costs while enhancing modeling efficiency and flexibility. This approach is particularly suitable for industrial environments with high uncertainty, temporal variability, and complex process dynamics.

The training pipeline for SSFM-TSA is depicted in Fig. \ref{fig3}(b). Similar to the first stage, the input processing remains consistent, where after obtaining token embeddings and positional embeddings, the two are summed and fed into the SSFM-TSA. Within the SSFM-TSA model, we introduce two serial TSAs with a minimal number of trainable parameters into each transformer layer. Following standard practices \cite{DBLP:journals/corr/abs-1902-00751}, these adapters are positioned respectively after the multi-head attention and the feed-forward layers. This insertion strategy ensures that while the model retains its original characteristics, it can flexibly adapt to the requirements of specific tasks. Except for TSAs, all other parameters in SSFM are frozen. TSA consists of three components: down project layer, a non-linear activation function GELU (gaussian error linear unit), and up project layer, forming a small trainable module with a bottleneck structure (as shown in Fig. \ref{fig5}(b)). Taking the TSA post-multi-head attention as an example, the down project first maps the high-dimensional hidden state vector ${{H}_{attn}}$ to a lower-dimensional space. This is followed by a transformation of the dimensionality reduction representation through GELU, and finally, the up project layer projects the low-dimensional representation back to the high-dimensional space of the original input. Furthermore, the TSAs employs a skip-connection to better preserve the original input information. This process can be represented by (9).
\begin{equation}
\resizebox{0.9\hsize}{!}{$\begin{aligned}
{{H}_{TSA}}={{W}_{up}}\left( GELU\left( {{W}_{down}}{{H}_{attn}}+{{b}_{down}} \right) \right)+{{b}_{up}}+{{H}_{attn}}
\end{aligned}$}
\end{equation}

Where ${{H}_{TSA}}$ denotes the output features of the TSA, with ${{W}_{down}}$ and ${{W}_{up}}$ representing the weight matrices of the dimensionality reduction and dimensionality increase layers, respectively. Meanwhile, ${{b}_{down}}$ and ${{b}_{up}}$ correspond to the bias vectors of these layers. $GELU\left( \cdot  \right)$ refers to the GELU activation function. Throughout the training process, all parameters within the SSFM are frozen, and we adapt to downstream tasks solely through fine-tuning the TSA and the task-specific linear layer. Upon completing the training, the SSFM-TSA possesses the capability to tackle multiple downstream soft-sensing modeling tasks. These tasks encompass the preprocessing stage of soft sensor modeling, including missing value imputation, and anomaly detection, as well as DDSS tasks during the soft sensor phase and tasks based on text knowledge embedding. Each task is equipped with its own adapter. During the inference stage, handling different tasks is achieved by assembling SSFM with respective TSAs, thereby implementing a universal industrial soft sensing modeling approach.

\subsection{Anomaly Detection}\label{4.5}

Industrial data anomaly detection is the first step in soft sensing modeling. By detecting and handling anomalies errors in the data can be removed or corrected, thereby enhancing the reliability of data analysis, modeling, and decision-making processes. However, anomaly labels in data collected from industrial environments are often inaccessible. Relying on domain experts for manual annotation of labels is not only labor-intensive but also inefficient and prone to errors. Therefore, we adopt a semi-supervised synthetic anomaly method to detect anomalies in sequences based on the similarity of sequence reconstruction. First, assuming sequence of the $s\text{-th}$ auxiliary variable is represented as ${{V}_{s}}$, if there are a total of $u+1$ samples, it can be expressed as follows:
\begin{equation}
{{V}_{s}}=\left[ {{x}_{s}}(0),\text{ }{{x}_{s}}(1),\text{ }{{x}_{s}}(2),\text{ }\ldots ,\text{ }{{x}_{s}}\left( u \right) \right]
\end{equation}

We treat all samples in ${{V}_{s}}$ as (pseudo-)positive labels. Additionally, we inject a specific type of anomaly (spikes anomaly, noise anomaly, scale anomaly) into random positions within ${{V}_{s}}$, treating these samples as (pseudo-)negative labels. The synthesis formula for spikes anomaly can be represented by follow:
\begin{equation}
{{A}_{Spike}}=m\cdot s,\text{ }m\sim Bernoulli(p),\text{ s}\sim \mathcal{N}\left( 0,{{\sigma }^{2}} \right)
\end{equation}

Where $p$ denotes the probability of a bernoulli variable $m$. When $m=1$, spikes noise is injected. In addition, we extract noise from $\mathcal{N}\left( 0,{{\sigma }^{2}} \right)$ to serve as noise anomaly and generate scale anomaly by scaling sequence segments through a scaling factor. Next, we divided the dataset into training, validation, and test sets. During the training phase, our objective was to reconstruct the sequence ${{X}_{s}}\left( t \right)$ of the $s\text{-th}$ auxiliary variable with minimal training epochs to obtain $X_{s}^{recon}\left( t \right)$, using MSE to optimize the training objective for anomaly detection adapter. The anomaly detection loss function ${{\mathcal{L}}_{AT}}$ can be represented by (12) and (13):
\begin{equation}
{{X}_{s}}\left( t \right)=[{{x}_{s}}\left( t \right),\text{ }{{x}_{s}}\left( t-1 \right),\text{ }\cdots \text{, }{{x}_{s}}\left( t-n+1 \right)]
\end{equation}
\begin{equation}
{{\mathcal{L}}_{AT}}=MSE\left( {{X}_{s}}\left( t \right),X_{s}^{recon}\left( t \right) \right)
\end{equation}

Finally, we proceed to reconstruct the auxiliary variable series on the test set and obtain the reconstruction errors between ${{X}_{s}}\left( t \right)$ and $X_{s}^{recon}\left( t \right)$. When the reconstruction error of a certain sample exceeds a predefined threshold, the sample is identified as an anomalous instance. This can be followed by utilizing missing value imputation to replace the identified anomaly.

\subsection{Missing Value Imputation}\label{4.6}

After annotating the anomalies, we integrate a missing value imputation adapter with the SSFM to train a downstream model for missing value imputation. Specifically, we first randomly select values from the input sequence $X\left( t \right)$ to mask, thereby obtaining ${{X}_{mask}}\left( t \right)$. ${{X}_{mask}}\left( t \right)$ replaces $X\left( t \right)$ as input into SSFM-TSA, and the reconstruction result $X_{mask}^{recon}\left( t \right)$ is derived through the final linear probing. $X\left( m \right)$ denotes the masked values in the input sequence, while $X_{mask}^{recon}\left( m \right)$ represents the reconstructed values at the masked locations. Ultimately, by minimizing the MSE, we compel $X_{mask}^{recon}\left( m \right)$ to increasingly converge with $X\left( m \right)$, culminating in the training of the missing value imputation adapter. The loss function ${{\mathcal{L}}_{MDI}}$ can be articulated as follow:
\begin{equation}
{{\mathcal{L}}_{MDI}}=MSE\left( X\left( m \right),X_{mask}^{recon}\left( m \right) \right)
\end{equation}

\subsection{LLM for Data-Driven Soft Sensor (LMM-DSS)}\label{LLM-DSS}

DDSS analyze the relationship between input and output process data of controlled objects to establish a black-box model, which possesses strong nonlinear processing capabilities and online correction abilities.

In this paper, we explore the potential of leveraging pre-trained LLM for addressing the classical approach of DDSS. Initially, we utilize the AVs time series $X\left( t \right)$ described in Section \ref{3.4} as the input for SSFM-TSA. After passing through the last transformer layer of the SSFM, we obtain the hidden state ${{H}^{L+K}}\in {{\mathbb{R}}^{n\times d}}$. Subsequently, ${{H}^{L+K}}$ is input into a linear layer adapted for the soft sensor task. The output dimension of this linear layer is 1, aimed at predicting the current time $t$ primary variable value $y_{t}^{predict}$, which can be represented by follow:
\begin{equation}
y_{t}^{predict}={{H}^{L+K}}W+b
\end{equation}
where $W\in {{\mathbb{R}}^{d\times 1}}$ and $b\in {{\mathbb{R}}^{1}}$ represent the weight matrix and bias vector of the linear layer, respectively. We also employ the MSE as the loss metric to compel the predicted values $y_{t}^{predict}$ to approximate the true measured values $y_{t}^{true}$.
\begin{equation}
{{\mathcal{L}}_{DSS}}=\frac{1}{c}\sum\limits_{t=1}^{c}{{{\left\| {{y}_{t}}-y_{t}^{\text{prediction}} \right\|}^{2}}}
\end{equation}

\subsection{The Proposed LLM-PSS}\label{3.h}

\begin{figure}[t]
\centering
\resizebox{\linewidth}{!}{
\includegraphics[width=3.5in]{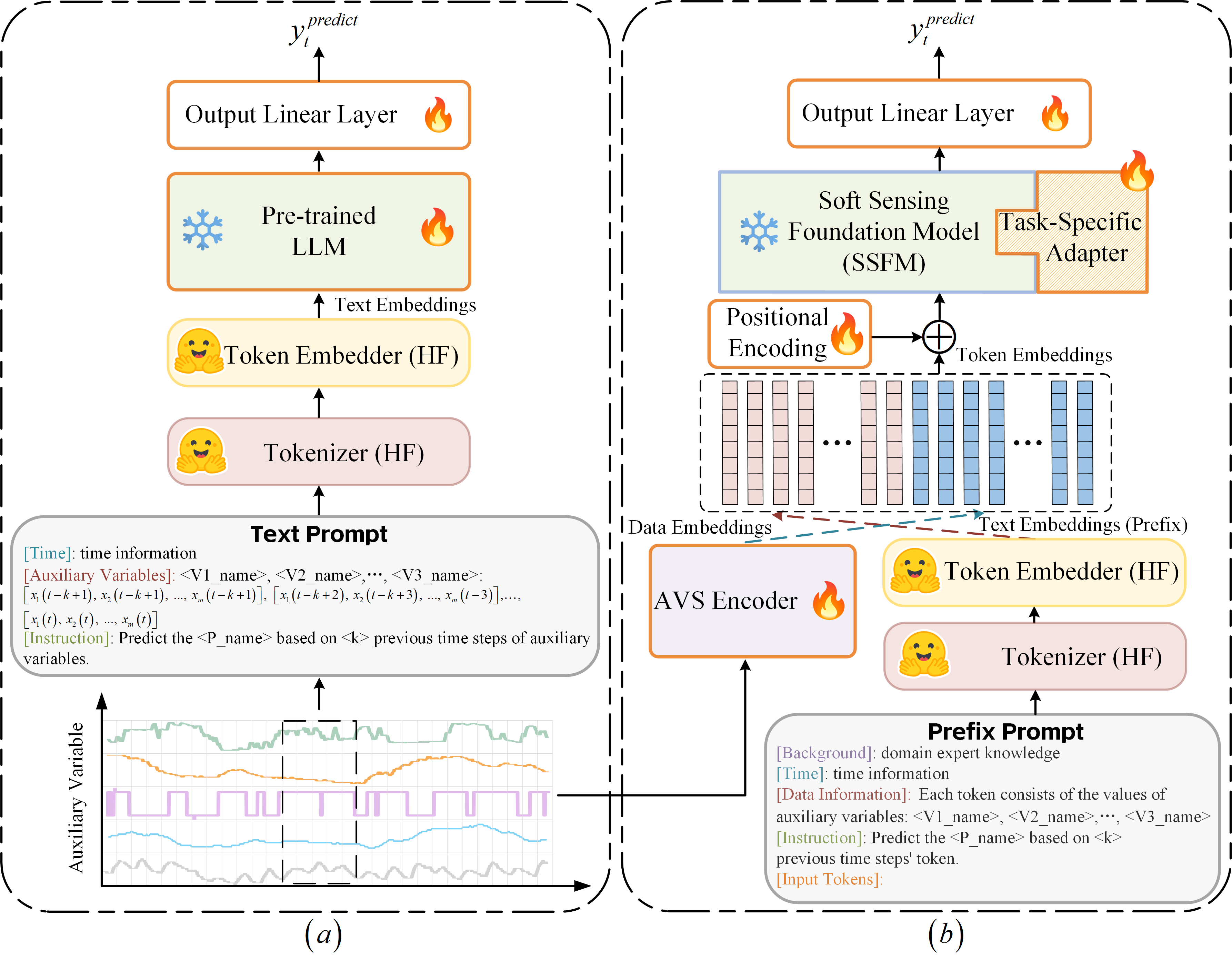}}
\caption{Framework of the proposed knowledge-embedded soft sensor (LLM-TKESS). (a) Proposed LLM-PSS. (b) Proposed LLM-PDSS.}
\label{fig6}
\end{figure}

Prompt learning represents a straightforward yet effective methodology for activating downstream tasks of pre-trained LLMs \cite{10.1145/3560815}. Prompt Learning transfigures the original input into a textual string prompt via templates, fed into the LLM to elicit the final output. Based on this, we designed the LLM-PSS, using this intuitive approach to demonstrate the capability of pre-trained LLMs in text and prompt-based soft sensing modeling, and to preliminarily explore the feasibility of soft sensing modeling through text.

The pipeline of LLM-PSS is illustrated in Fig. \ref{fig6}(a). Initially, we transform the input sequence of AVs into text-based prompt inputs through templates. We selected the four most crucial aspects for constructing the text prompt template for the soft sensing dataset: background, time, AVs, and instruction. The background provides an introduction to the core principles of the industrial domain; Time annotates the current time information; The AVs section is populated with the names of all AVs, from V1\_name to Vm\_name; and the instruction offers a language description to guide the LLM in generating specific types of outputs. After obtaining the text prompt input, it is fed into the standard pre-trained tokenizer (HF) and embedder (HF) provided by HuggingFace (HF) to obtain the text embeddings. Subsequently, the text input is passed into the pre-trained LLM for fine-tuning, similar to the procedure in Section \ref{3.3}. The current time's output $y_{t}^{predict}$ is then obtained through a linear layer, with the loss function for this training process being the same as in Section \ref{LLM-DSS}.

\subsection{The Proposed LLM-PDSS}

Through the experiments in Section \ref{4.6}, we demonstrated that LLM-PSS is capable of performing soft sensor modeling through text. Although this method is straightforward and intuitive, it is challenging for LLMs to track auxiliary variable data for reasoning. Currently, various prompt learning methods have shown promising results by using prompt-as-prefix to integrate multimodal information and enhance the model's reasoning capabilities \cite{10.5555/3540261.3540277,jin2024timellm}. Based on this, we further propose a second type of TKESS, LLM-PDSS, which incorporates a prefix prompt with industrial background information and temporal information into the DDSS described in Section \ref{LLM-DSS}.

The architecture of LLM-PDSS is illustrated in Fig. \ref{fig6}(b). Initially, Prefix prompts containing background, temporal, and data information, along with instructions, are sequentially input into the tokenizer (HF) and embedder (HF) to generate text embeddings. It is crucial to note that, unlike text prompts, prefix prompts do not encompass the specific values of AVs. Next, the input sequence is fed into the AVS Encoder to obtain data embeddings. Text embeddings are used as prefixes to concatenate with data embeddings, followed by the addition of positional encoding. Subsequently, the concatenated result is fed into SSFM-TSA and the output linear layer for downstream task training. The training process and loss function are the same as described in Section in \ref{LLM-DSS}.

\begin{figure}[t]
\centering
\includegraphics[width=3.3in]{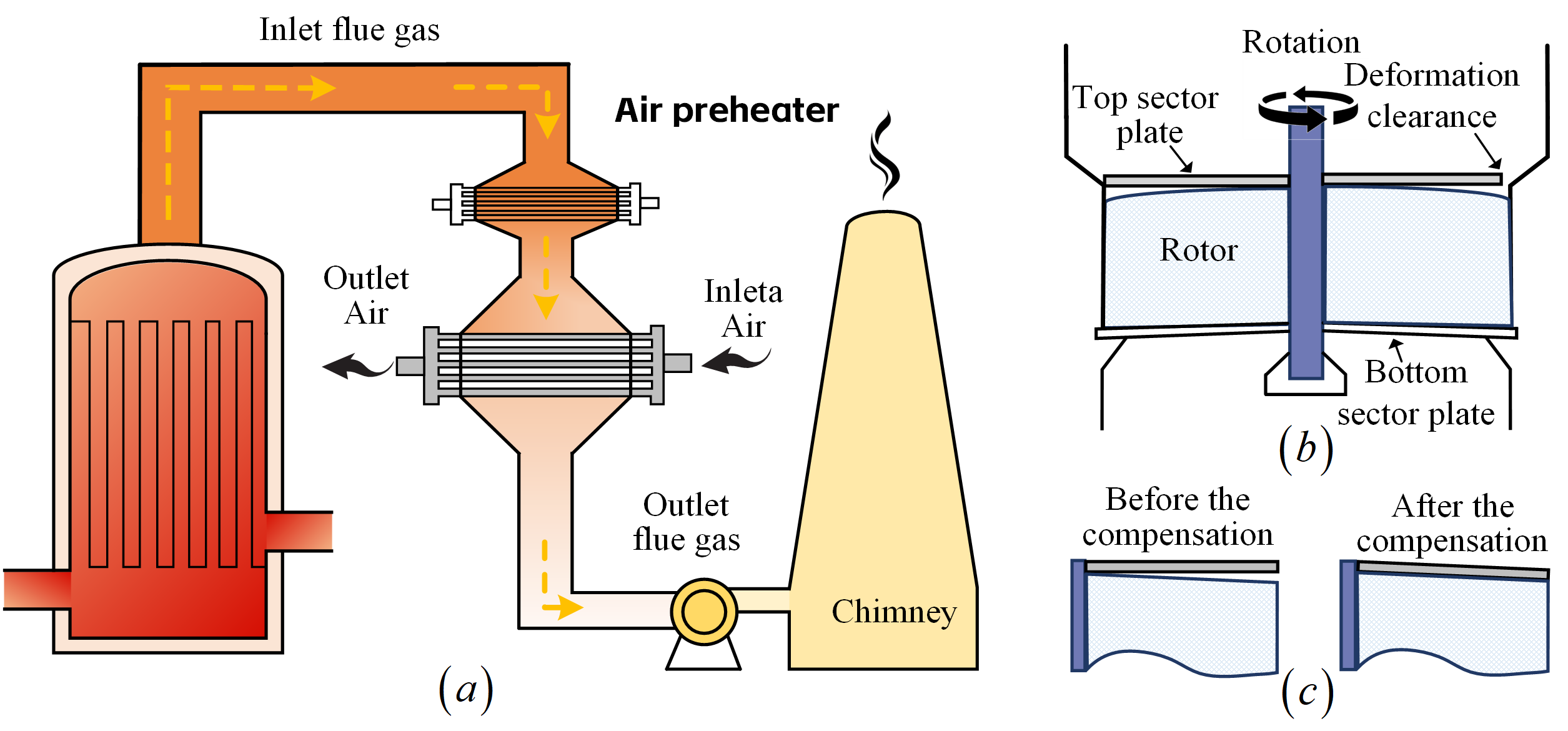}
\caption{Schematic diagram of the air preheater. (a) Heat exchange principle of the air preheater. (b) Schematic diagram of rotor thermal deformation. (c) Schematic diagram of clearance compensation.}
\label{fig7}
\end{figure}

\section{Experiment details}

\subsection{Case Descriptions}\label{4.1}

As shown in Fig. \ref{fig7}(a), the rotary air preheater utilizes the heat from the waste flue gas of boilers to preheat the combustion-supporting air, thereby achieving exhaust recovery and utilization. This principle of heat exchange effectively improves energy efficiency and is widely used in thermal power plants. However, air preheaters often experience air leakage issues due to their structural characteristics. This is caused by uneven heating of the rotor during the heat exchange process, resulting in complex thermal stresses inside the rotor that lead to mushroom-shaped thermal deformation \cite{WANG2018112}. As shown in Fig. \ref{fig7}(b), the thermal deformation of the rotor increases the clearance between the top sector plate and the rotor, leading to air leakage. This phenomenon results in significant energy waste and economic losses \cite{10197236}. 

Based on issues, it is crucial to detect rotor thermal deformation using sensors, and then utilize the clearance control system to adjust the height of the top sector plate for compensation (as shown in Fig. \ref{fig7}(c)), aiming to minimize air leakage. However, the air preheater operates in high-temperature environments where the flue gas contains a significant amount of dust. This situation makes traditional hardware sensors highly susceptible to damage and unable to operate safely and reliably under these conditions \cite{WANG2018112}. Therefore, constructing a soft sensor based on various temperature process variables easily obtained by sensors holds significant research value for stable and accurate measurement of rotor thermal deformation.

\begin{figure}[t]
\centering

\includegraphics[width=3in]{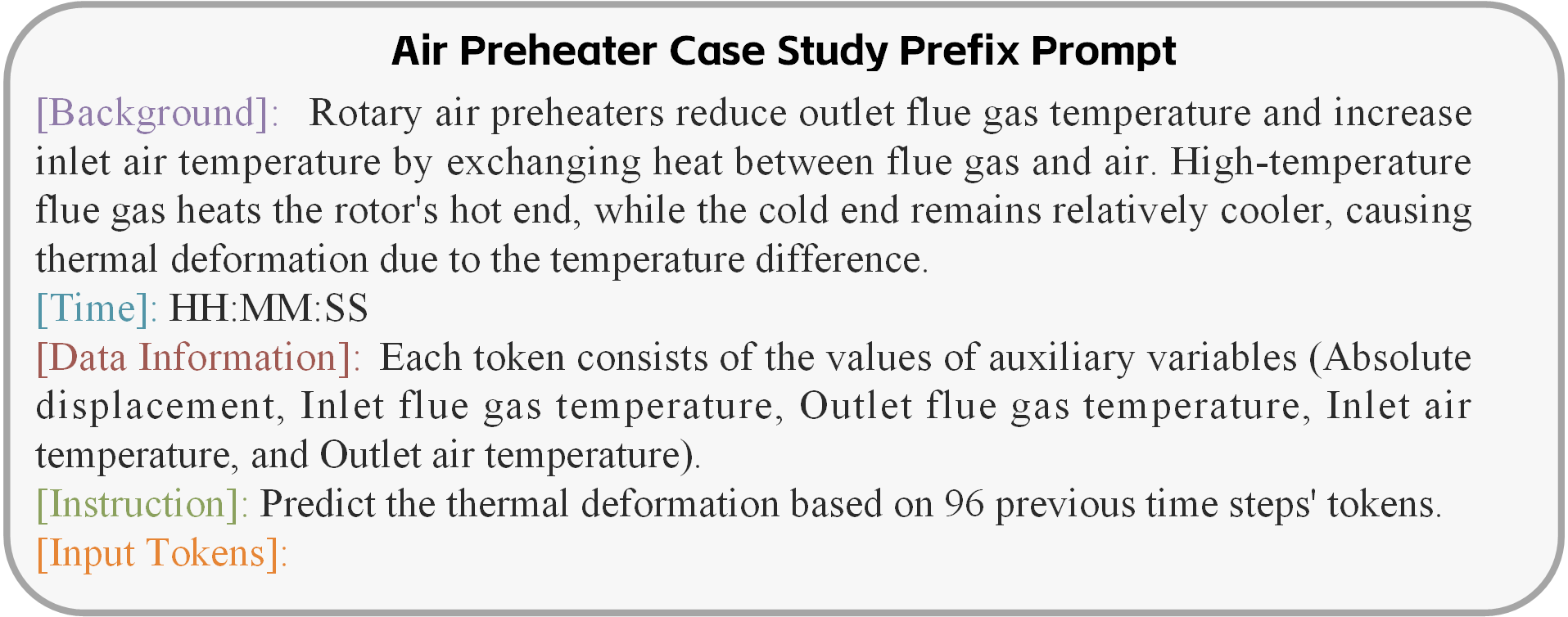}
\caption{Air preheater case study prefix prompt.}
\label{fig8}

\end{figure}

Based on the industrial requirements, we utilize the proposed general industrial soft sensing architecture, LLM-TKESS, to implement soft sensing modeling for this industrial process. The rotor thermal dataset used in this study was collected from a thermal power plant in Shandong Province, China. The AVs include the inlet temperature of the gas flue, the outlet temperature of the gas flue, the inlet temperature of the air duct, the outlet temperature of the air, and the absolute displacement (denoted as ${{v}_{1}}$,${{v}_{2}}$,${{v}_{3}}$,${{v}_{4}}$,${{v}_{5}}$). The primary variable of interest is the rotor thermal deformation. Additionally, based on the aforementioned principles, we designed an air preheater case study prefix prompt as auxiliary input information for LLM-PDSS, with detailed specifics shown in Fig. \ref{fig8}.

\subsection{Parameter Setting}
In this paper, all experiments were conducted on the rotary air preheater dataset introduced in Section \ref{4.1}. The dataset comprises a total of 23,000 samples, with a sampling interval of 65 seconds between samples. We partitioned the dataset into training, validation, and test sets in a ratio of 7:1.5:1.5.

Our experiments were implemented in PyTorch and conducted on RTX 3090 (24GB) GPUs. To reduce the impact of randomness, all models were trained three times, and the average of the three evaluation metrics was taken as the final performance measure. We utilized the first 6 layers of GPT-2 \cite{Radford2019LanguageMA} as our pretrained large model, sourced from HuggingFace. Adam optimizer \cite{DBLP:journals/corr/KingmaB14} with a learning rate of 0.0001 and a batch size of 64 were used in our experiments. the configurations and descriptions of the model parameters at each stage are detailed in Table \ref{table1}.

\begin{table}[t]
\caption{\textbf{An overview of the experimental parameter configuration for LLM-TKESS}}%标题
\label{table1}
\centering%把表居中
\begin{tabular}{lclc}
\toprule%第一道横线
Stage& Parameter &Description & Value\\
\midrule%第二道横线 
AVS Encoder&$k$&1D convolution kernel size&$3\times 3$\\
  &$p$&Padding&1\\
  &$s$&Stride&1\\
  &$d$&Embedding dimension&768\\
SSFM&${{r}_{LoRA}}$&LoRA dimension&4\\
  &$\alpha $&LoRA scaling factor&32\\
  &${{p}_{LoRa}}$&LoRA dropout&0.1\\
   &$K$&LN fine-tuning layer&4\\
   &$L$&LN LoRA fine-tuning layer&2\\
SSFM-TSA&${{r}_{Adapter}}$&Adapter dimension&32\\
    &${{p}_{Adapter}}$&Adapter dropout&0.1\\
   
\bottomrule%第三道横线

\end{tabular}

\end{table}

\subsection{Evaluation Criterion}
In this study, the mean absolute error (MAE), coefficient of determination (${{R}^{\text{2}}}$), root mean square error (RMSE), and mean absolute percentage error (MAPE), as shown in (17)-(21), are used as evaluation criteria to assess the performance of LLM-TKESS.
\begin{equation}
\text{MAE}=\frac{1}{c}\sum\limits_{t=1}^{c}{\left| {{y}_{t}}-y_{t}^{\text{prediction}} \right|}
\end{equation}
\begin{equation}
{{R}^{2}}=1-\frac{\sum\nolimits_{t=1}^{c}{{{\left( {{y}_{t}}-y_{t}^{\text{prediction}} \right)}^{2}}}}{\sum\nolimits_{t=1}^{c}{{{\left( {{y}_{t}}-\overline{y} \right)}^{2}}}}
\end{equation}
\begin{equation}
\text{RMSE}=\sqrt{\frac{1}{c}\sum\limits_{t=1}^{c}{{{\left( {{y}_{t}}-y_{t}^{\text{prediction}} \right)}^{2}}}}
\end{equation}
\begin{equation}
\text{MAPE =}\frac{1}{c}\left| \sum\limits_{t=1}^{c}{\frac{{{y}_{t}}-y_{t}^{\text{prediction}}}{{{y}_{t}}}} \right|\times 100\%
\end{equation}
\begin{equation}
\text{SMAPE =}\frac{1}{c}\sum\limits_{t=1}^{c}{\frac{\left| {{y}_{t}}-y_{t}^{\text{prediction}} \right|}{\left( \left| {{y}_{t}} \right|+\left| y_{t}^{\text{prediction}} \right| \right)/2}}\times 100\%
\end{equation}

Where $c$ represents the number of samples in the test dataset, and $\bar{y}$ denotes the mean value of all data labels in the test set. MAE is used to gauge the average magnitude of the absolute errors between the predicted and actual values, while RMSE measures the square root of the mean squared error between these values. The closer the MAE and RMSE values are to 0, the closer the predictions are to the actual values. The ${{R}^{2}}$ score ranges from 0 to 1, with higher values indicating smaller residuals between the predicted and actual values and, therefore, a better model fit to the data. MAPE measures the error between predicted and actual values as a percentage, while SMAPE quantifies the relative error percentage between predicted and actual values. Smaller values of both metrics indicate higher prediction accuracy.

\subsection{Experimental Results of LMM-DSS and LLM-PDSS}

To validate the superior modeling capabilities of the proposed LLM-DSS and LLM-PDSS based on LLM, we selected several state-of-the-art soft sensors for comparative performance evaluation on the test dataset. These include: (1) Semi-supervised variational autoencoders for regression (SSVAER) improved with variational autoencoders (VAE) \cite{zhuang2022semisupervised}; (2) Kolmogorov-arnold networks (KAN) \cite{liu2024kan}, currently the latest and most promising alternative to MLP; (3) Semi-supervised stacked autoencoder (SSAE) enhanced with stacked autoencoders \cite{YUAN2020115509}; (4) Variable attention-based long short-term memory (VA-LSTM) network combined with attention mechanisms \cite{https://doi.org/10.1002/cjce.23665}; (5) Spatiotemporal attention-based long short-term memory (STA-LSTM) \cite{9062588}; In addition to these advanced deep learning models, we also included traditional machine learning regression algorithms for comprehensive comparison: (6) Support vector regression (SVR) optimized through grid search for optimal parameters; (7) Stacking regressor (STACKINGR) using k-nearest neighbors regression, ridge regression as base learners, and multi-layer perceptron as meta learner.

\begin{table}[t]
\caption{\textbf{Comparative results of various soft sensor }}%标题
\label{table2}
\centering%把表居中

\begin{tabular}{lcccc}
\toprule%第一道横线
Methods  & MAE $\downarrow$ &  RMSE $\downarrow$&SMAPE $\downarrow$ & ${{R}^{2}}$ $\uparrow$ \\
\midrule%第二道横线 
SVR &0.0492&0.0616&9.1166&0.8398\\
STACKINGR &0.0434&0.0578&7.6446&0.8589 \\
SSVAER &0.0406&0.0546&7.6675&0.8793\\
KAN &0.0464&0.0578&8.8543&0.8543\\
VA-LSTM &0.0392&0.0507&7.1937&0.8915\\
SSAE  &0.0390&0.0501&7.3335&0.8991\\
STA-LSTM &0.0376&0.0508&6.7246&0.8909\\
LLM-DSS (ours) &0.0335&0.0470&5.9471&0.9063\\
\textbf{LLM-PDSS (ours)}&\textbf{0.0329}&\textbf{0.0456}&\textbf{5.9380}&\textbf{0.9121}\\
\bottomrule%第三道横线
\end{tabular}
\end{table}

\begin{figure}[t]
\centering
\resizebox{\linewidth}{!}{
\includegraphics[width=3.5in]{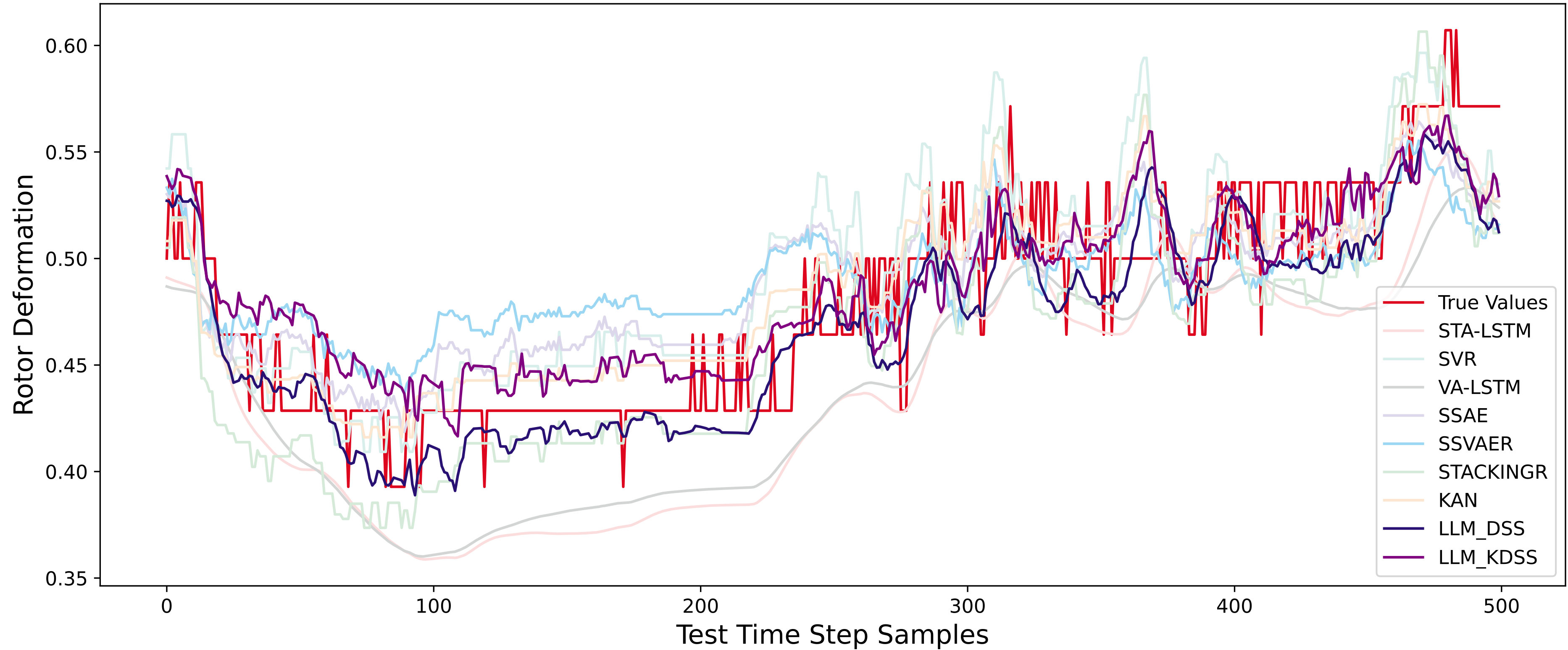}}
\caption{Prediction curves of rotor deformation by 9 models on continuous test samples.}
\label{fig9}
\end{figure}

We initially trained the models mentioned above using the entire training dataset, followed by validation on the test set. The comparative results of the 9 models' evaluation metrics are shown in Table \ref{table2}. It can be observed that the proposed LLM-PDSS and LLM-DSS models outperformed the others, achieving the highest and second-highest accuracy rates, respectively. For the RMSE, LLM-PDSS showed improvements of 25.97\%, 21.11\%, 16.48\%, 21.11\%, 10.06\%, 8.98\%, and 10.24\% compared to SVR, STACKINGR, SSVAER, KAN, VA-LSTM, SSAE, and STA-LSTM, respectively.

\begin{figure}[t]
\centering

\includegraphics[width=3.3in]{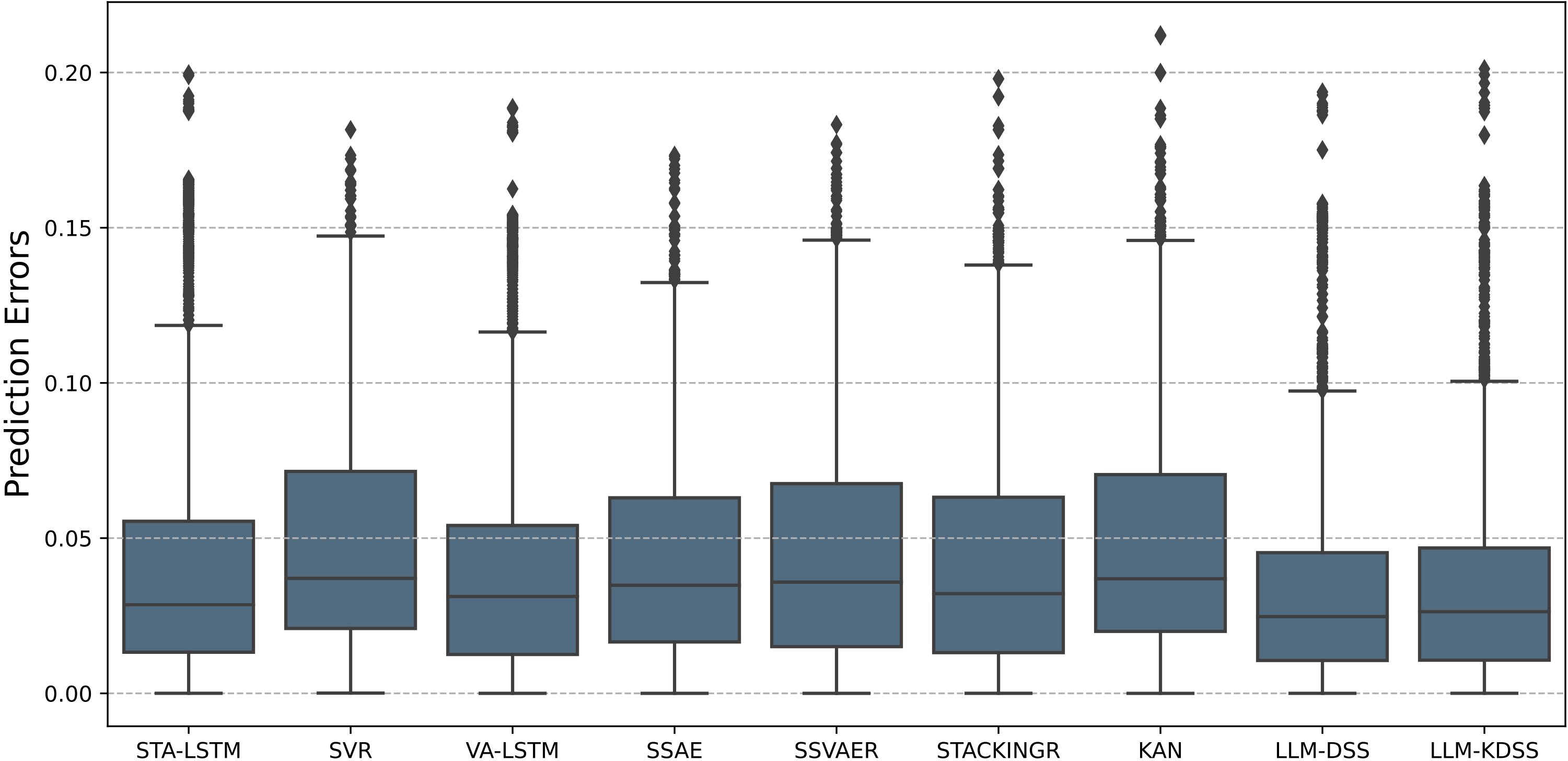}
\caption{Boxplots of errors for different models.}
\label{fig10}
\end{figure}

Fig. \ref{fig9} illustrates the testing results of the evaluated models on a continuous sequence of time-series samples. To facilitate observation, the vertical axis has been adjusted to the range of 0.3 to 0.65. The red curve represents the true rotor deformation. As shown in Fig. \ref{fig9}, LLM-PDSS and LLM-DSS exhibit a closer alignment with the actual rotor deformation trend compared to the other models, yielding predictions with smaller errors. This validates the conclusions derived from Table \ref{table2}. In addition, we have plotted the error box plots for all models, as shown in Fig. \ref{fig10}. The error comparison clearly demonstrates that the prediction results of LLM-DSS and LLM-PDSS are more densely distributed and are centered around zero. Based on these experiments, it is evident that the proposed LLM-TKESS, with its robust multimodal and temporal processing capabilities, exhibits superior predictive potential and performance compared to current soft sensing methods.

\subsection{Few-Shot Soft Sensing Experimental Results}

LLMs have undergone extensive and dense training on high-quality, large-scale corpora, which endows them with a rich repository of general world knowledge. This comprehensive training has resulted in LLMs exhibiting a highly potent cross-modal emergent capability that is not observed in other models. This cross-modal emergent ability is particularly evident in the context of handling few-shot tasks and has been substantiated across various few-shot downstream tasks. Moreover, research indicates that the underlying layers of large language models are engaged in some general processes that are not necessarily related to text data \cite{zhou2023fits}. To validate the emergent capabilities of LLMs and to assess whether their intrinsic propensity for sequence data reasoning can benefit the resolution of few-shot tasks in soft sensing, we retrained the aforementioned models using only 10\% of the training dataset samples. These models were then evaluated on the test set for comparison. The comparison results are presented in Table \ref{table3}.

\begin{table}[t]
\caption{\textbf{Comparison of few-shot soft sensing prediction results on 10\% training data}}%标题
\label{table3}
\centering%把表居中

\begin{tabular}{lcccc}
\toprule%第一道横线
Methods & MAE $\downarrow$ &  RMSE $\downarrow$&SMAPE $\downarrow$ & ${{R}^{2}}$ $\uparrow$ \\
\midrule%第二道横线 
SVR &0.0526&0.0756&9.4383&0.7523\\
STACKINGR &0.0991&0.1312&17.7636&0.2731 \\
SSVAER &0.0521&0.0775&8.4502&0.7482\\
KAN &0.1059&0.1404&21.0341&0.1131\\
VA-LSTM &0.1912&0.2309&38.2403&-1.2480\\
SSAE  &0.0514&0.0708&9.1957&0.7947\\
STA-LSTM &0.1945&0.2343&39.4487&-1.3574\\
LLM-DSS (ours) &0.0405&0.0530&8.1963&0.8814\\
\textbf{LLM-PDSS (ours)}&\textbf{0.0381}&\textbf{0.0485}&\textbf{7.3724}&\textbf{0.9008}\\
\bottomrule%第三道横线
\end{tabular}

\end{table}

From the results shown in the table, it can be observed that LLM-PDSS and LLM-DSS exhibit the capability to identify complex data patterns from limited data. In contrast, other methods did not demonstrate few-shot modeling capabilities, with the VA-LSTM and STA-LSTM soft sensors based on LSTM even showing negative values. Among the baseline methods, the SSAE model exhibited the best few-shot performance. However, LLM-PDSS improved the RMSE by 52.94\% compared to SSAE. This indicates that LLM-TKESS demonstrates a robust predictive capability for adaptive soft sensing in few-shot scenarios, which is unprecedented in previous models. This robustness is particularly significant for addressing complex industrial environments where data samples are difficult to collect. It is noteworthy that under well-sampled training conditions as shown in Table \ref{table2}, the evaluation results of LLM-DSS and LLM-PDSS for data-driven soft sensing modeling methods are relatively similar. However, in the few-shot scenario, the evaluation results of LLM-PDSS are significantly superior to those of LLM-DSS. This confirms that incorporating domain expert knowledge and relevant instructions can significantly enhance the model's prediction accuracy in small sample contexts.

\begin{figure}[!t]
\centering
\subfloat[]{\includegraphics[width=1.74in]{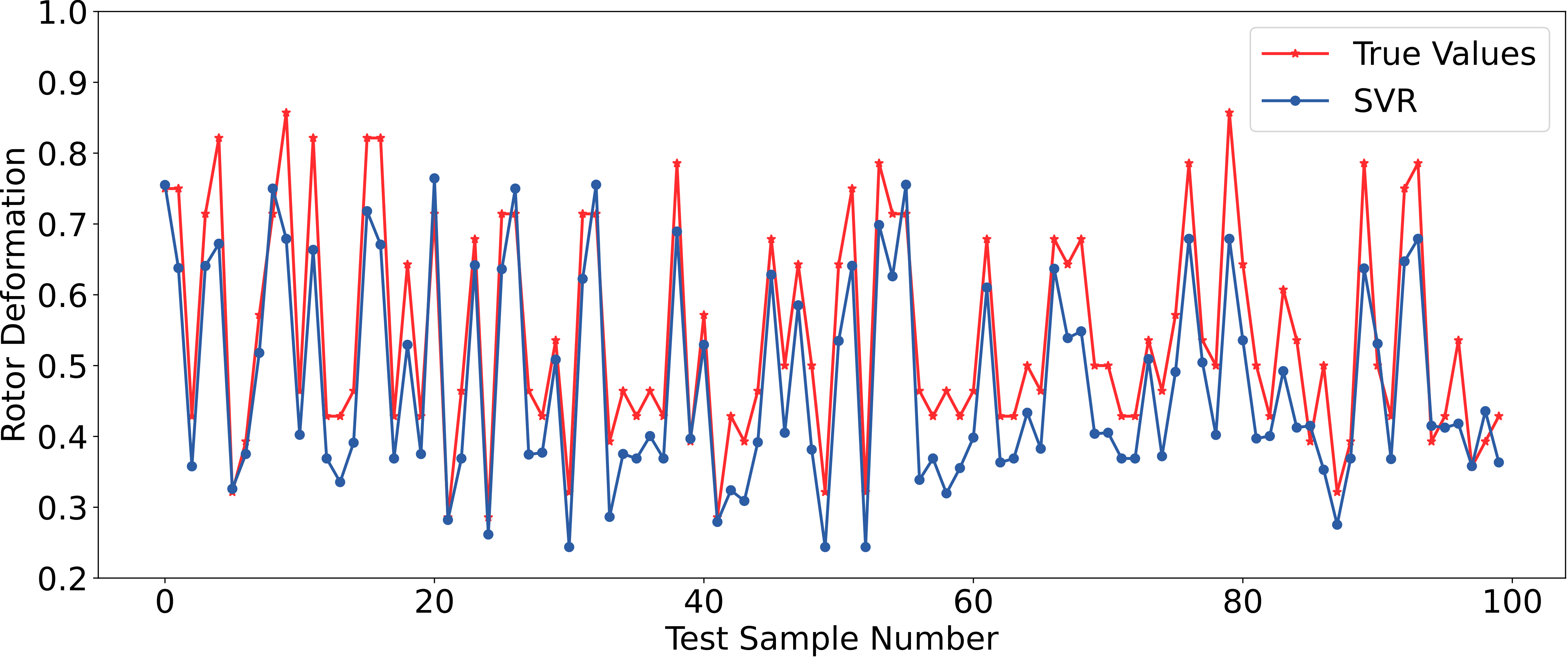}%
\label{fig_11_first_case}}
\hfil
\subfloat[]{\includegraphics[width=1.74in]{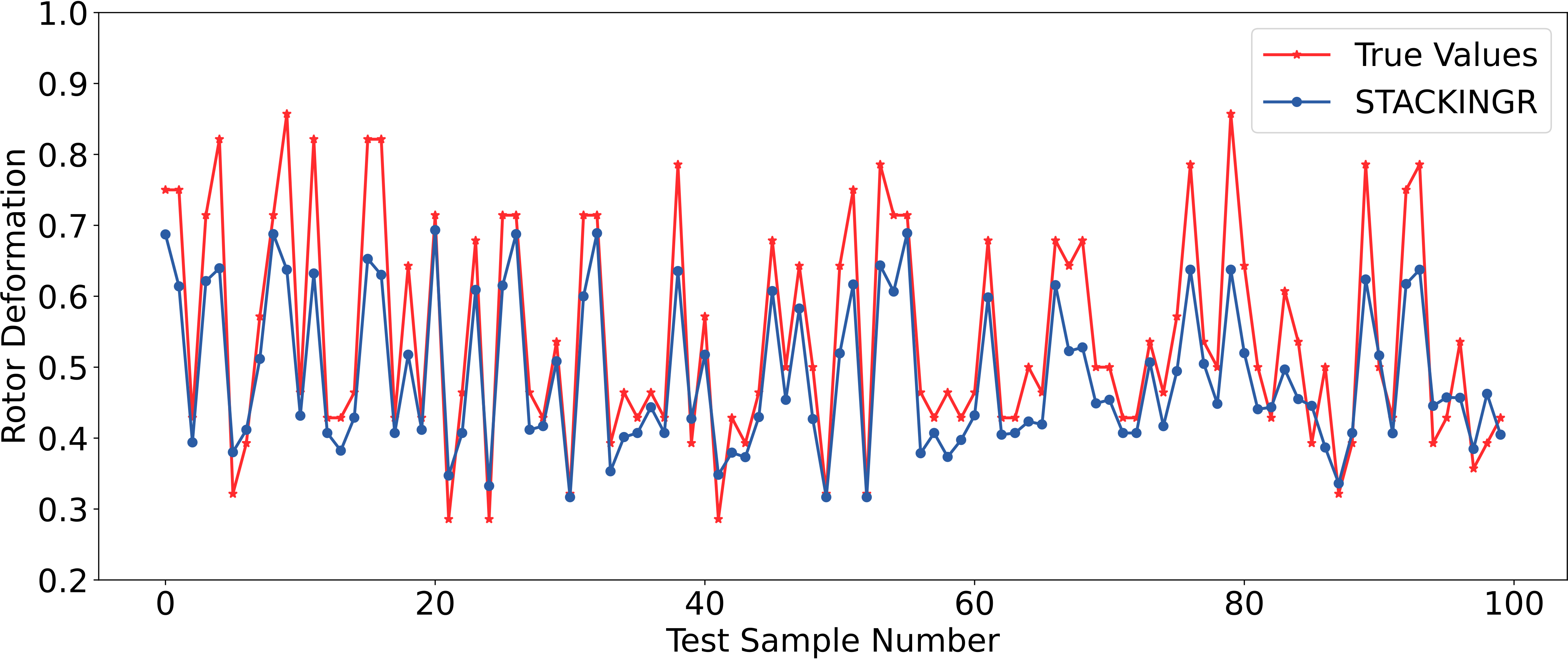}%
\label{fig_11_second_case}}

\subfloat[]{\includegraphics[width=1.74in]{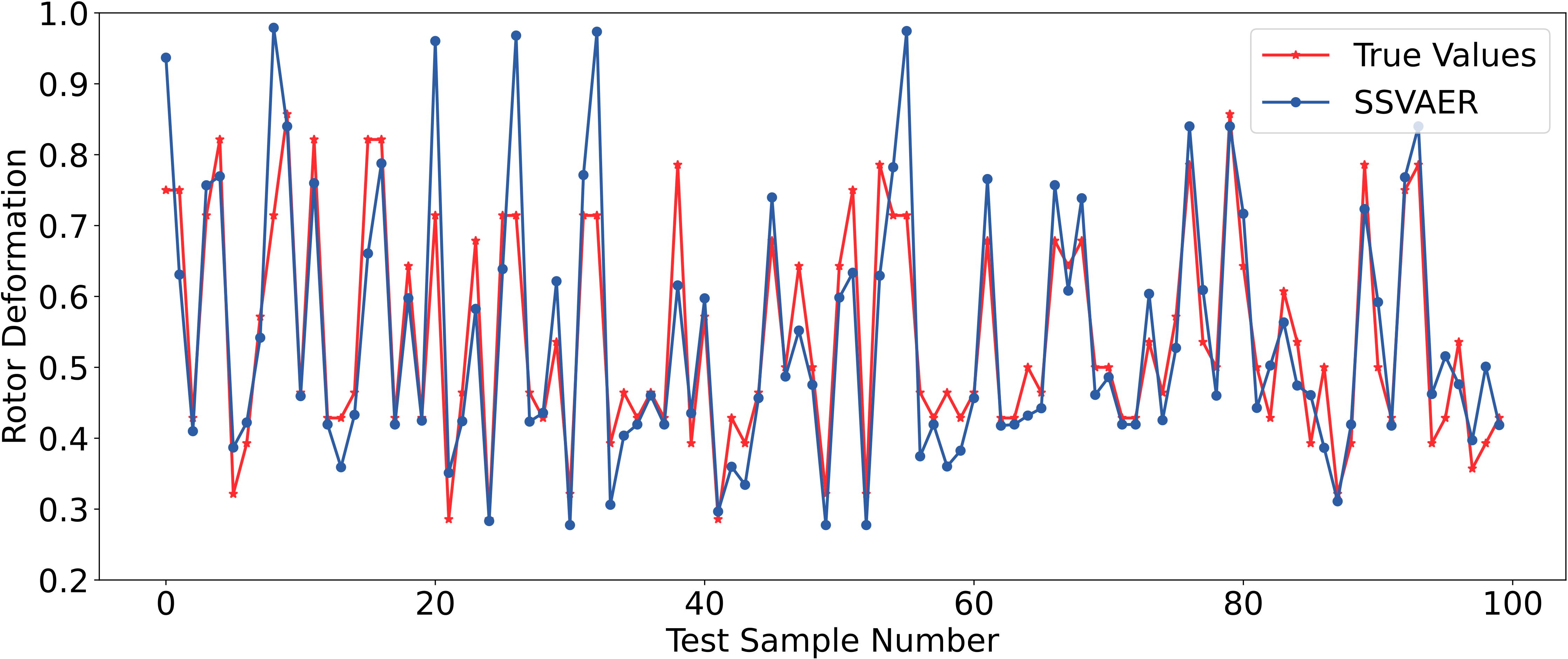}%
\label{fig_11_third_case}}
\hfil
\subfloat[]{\includegraphics[width=1.74in]{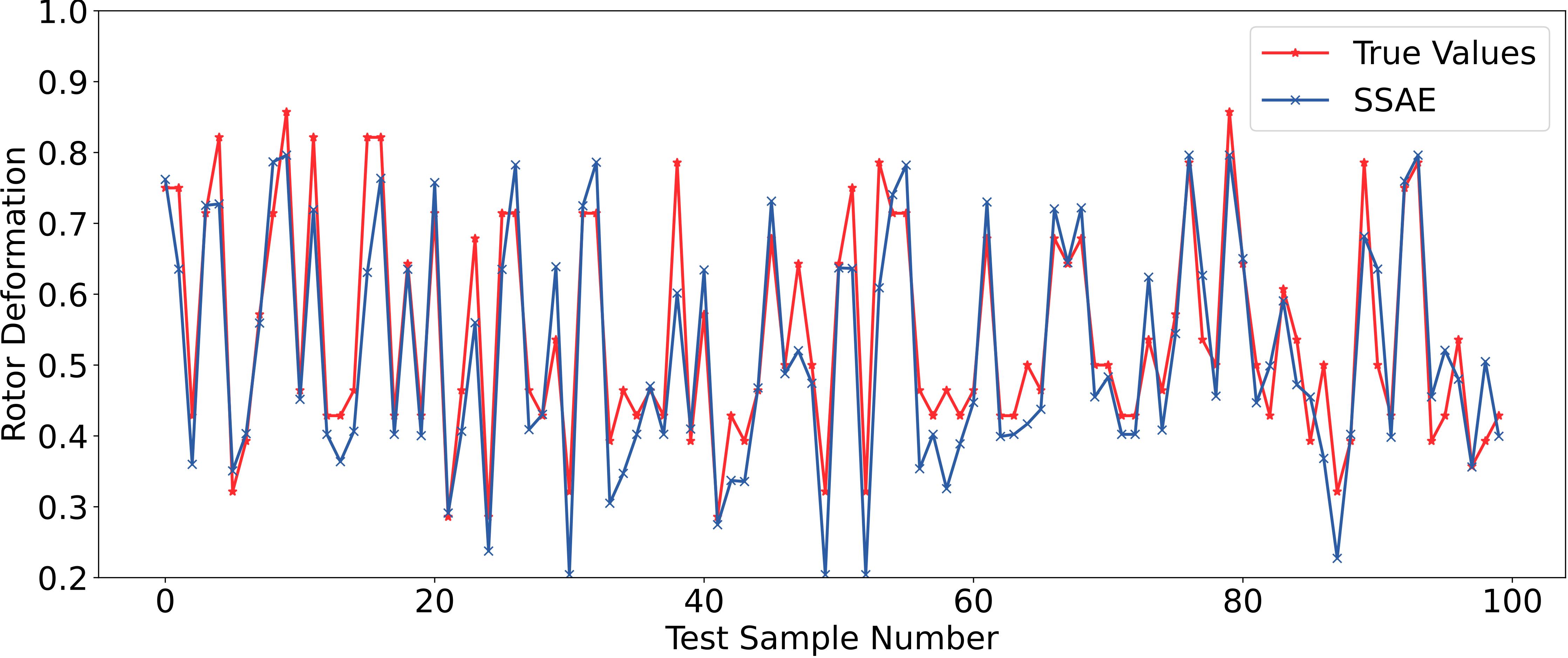}%
\label{fig_11_fourth_case}}

\subfloat[]{\includegraphics[width=1.74in]{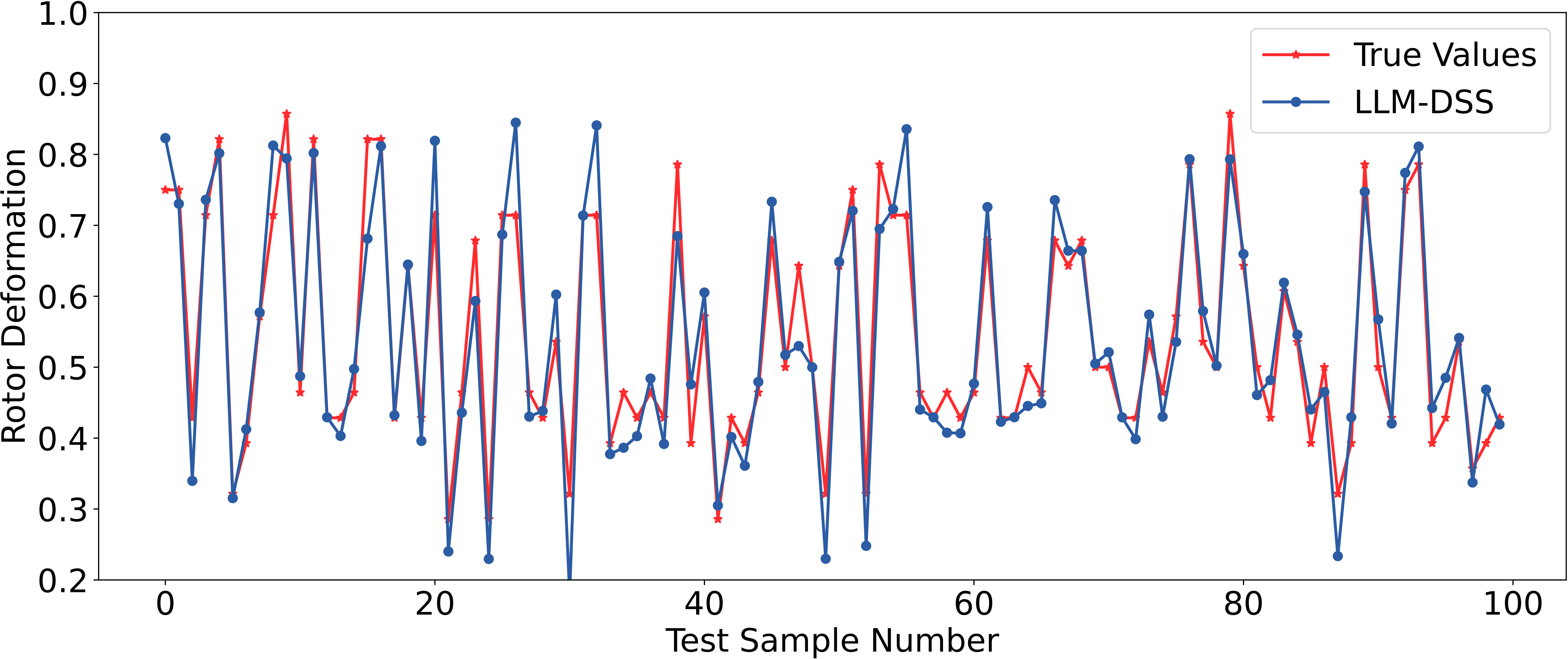}%
\label{fig_11_fifth_case}}
\hfil
\subfloat[]{\includegraphics[width=1.74in]{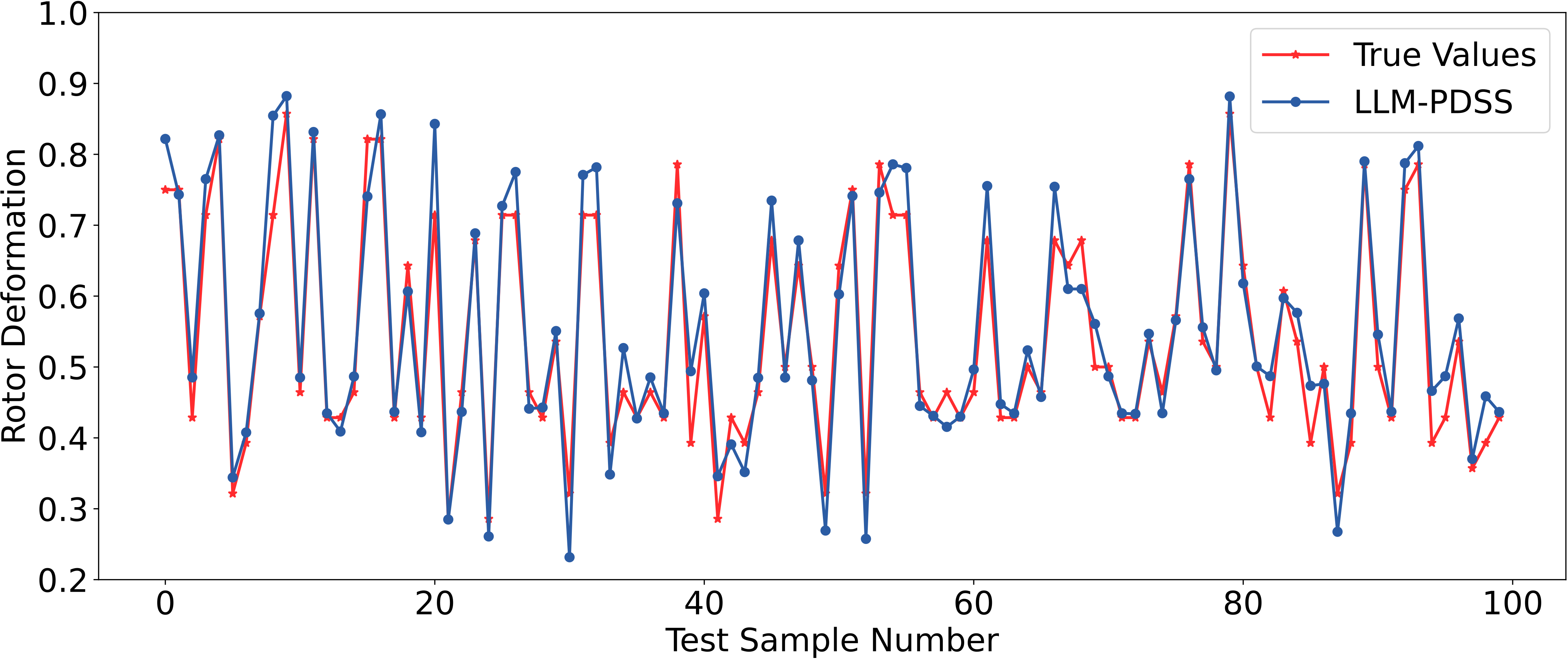}%
\label{fig_11_sixth_case}}

\caption{Comparison of the prediction results of 6 different soft sensing methods on 100 random points under few-shot conditions. (a) SVR. (b) STACKINGR. (c) SSVAER. (d) SSAE. (e) LLM-DSS. (f) LLM-PDSS.}

\label{fig11}
\end{figure}

We selected the six models with the smallest prediction errors (SVR, STACKINGR, SSVAER, SSAE, LLM-DSS, and LLM-PDSS) for a visual analysis of their prediction results, as shown in Fig. \ref{fig11}. Compared to the other models, it is evident that LLM-DSS and LLM-PDSS exhibit a superior ability to accurately fit the actual rotor deformation values. This is attributed to the inherent representation learning capabilities developed during their pre-training processes. Through cross-modal knowledge transfer, these models demonstrate an enhanced ability to perform non-linear fitting even in few-shot conditions.

\subsection{Results of LLM-PSS}\label{4.6}

LLM-PSS, by facilitating prediction access through sentence-based prompts, significantly broadens the potential for soft sensing modeling and enhances the interpretability of results. This approach offers a more intuitive and user-friendly method for soft sensing. We encoded the training dataset into prompt text based on the rotary air preheater data and the template proposed in Section \ref{3.h}. Each prompt text training sample encompassed an input time window of 20. It is important to note that the input to LLM-PSS is sentence-based. To accommodate practical usability and user-friendly requirements, the values of the AVs in the prompt text were not normalized. Consequently, the predicted values of the primary variable did not require any post-processing. Moreover, because the input alignment between the autoregressive fine-tuning phase of the SSFM and text input was inconsistent, we directly used a 6-layer pre-trained LLM for full parameter fine-tuning of the LLM-PSS.

\begin{table}[t]
\caption{\textbf{Performance comparison of LLM-PSS with other methods}}%标题
\label{table4}
\centering%把表居中

\setlength{\tabcolsep}{7mm}{
\begin{tabular}{lcc}
\toprule%第一道横线
Methods& MAE $\downarrow$ &  RMSE $\downarrow$\\
\midrule%第二道横线 
SVR &1.4249&1.8419\\
STACKINGR &1.3585&1.7142 \\
SSVAER &3.9266&7.8673\\
KAN &1.4170&1.8201\\
VA-LSTM &3.5767&4.3755\\
SSAE  &3.9358&7.9890\\
STA-LSTM &3.5928&4.4079\\
\textbf{LLM-PSS (ours)}&\textbf{1.2984}&\textbf{1.6123}\\
\bottomrule%第三道横线
\end{tabular}}
\end{table}

We conducted a performance comparison between the LLM-PSS and a selection of baseline models. The results for RMSE and MAE are presented in Table \ref{table4}. It is evident that data normalization has a significant impact on the performance of machine learning and deep learning models. In the baseline methods, deep learning models were particularly sensitive to changes in the scale of input data. The lack of data normalization led to difficulties in weight optimization during training, and in some cases, the models failed to converge. 

\begin{figure}[t]
\centering

\includegraphics[width=2.6in]{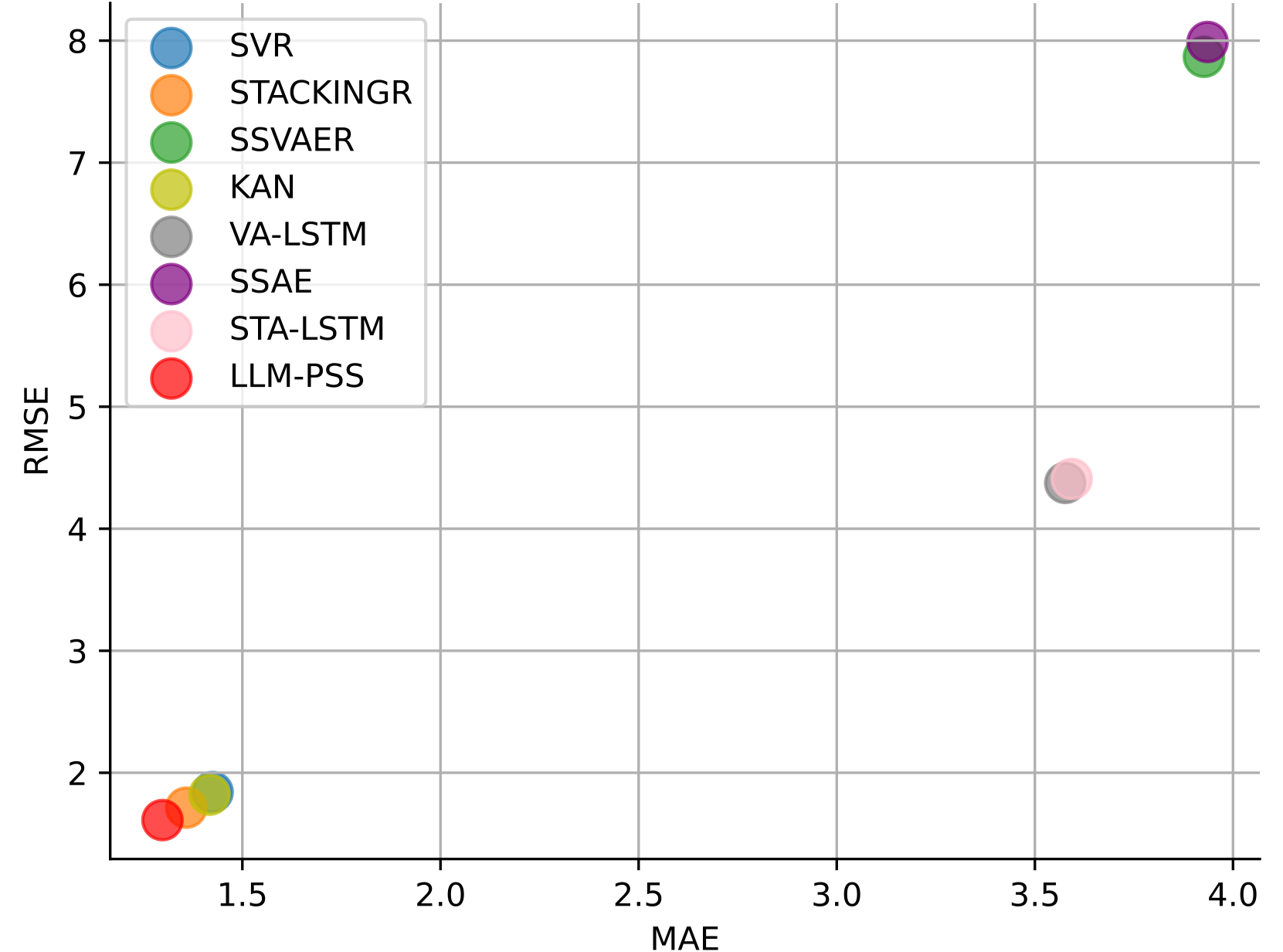}
\caption{The error comparison with the RMSE and MAE.}
\label{fig17}
\end{figure}

LLM-PSS inherently relies on text-based inputs, demonstrating significant robustness to changes in data scale. As illustrated in Fig. \ref{fig17}, LLM-PSS consistently achieves the lowest MAE and RMSE among the eight methods compared, underscoring the potential of text data, coupled with the sequential processing capabilities of LLM, to excel in soft sensing tasks. 

\begin{figure}[t]
\centering

\includegraphics[width=2.6in]{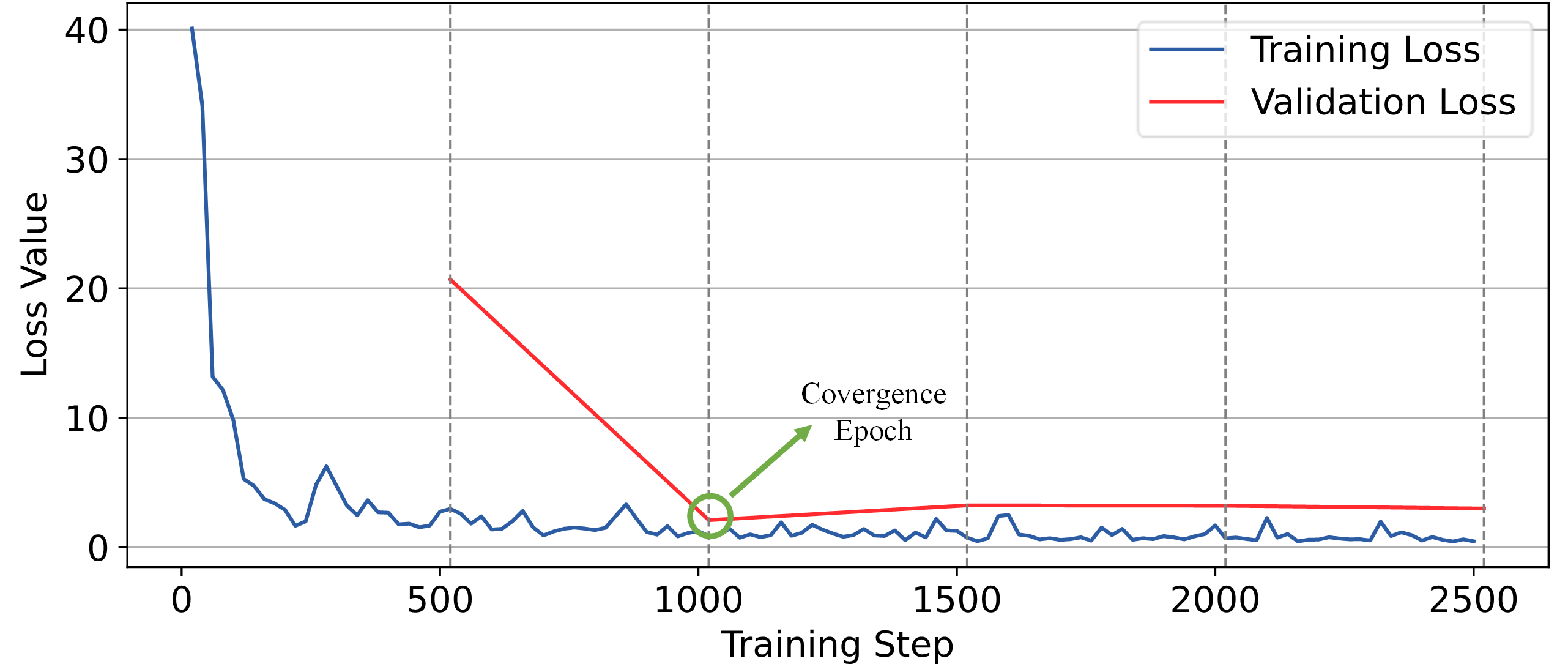}
\caption{ The training process of LLM-PSS.}
\label{fig18}
\end{figure}

A key aspect of evaluating the effectiveness of LLM-PSS in modeling the primary variable lies in the model’s convergence behavior during training. Fig. \ref{fig18} presents the training and validation error curves of LLM-PSS. The training error is recorded every 20 steps, while the validation error is logged at each epoch, represented by the dashed lines in the figure. As training progresses, the training error of LLM-PSS continuously decreases, indicating good convergence. The fine-tuning of LLM-PSS, leveraging a pre-trained LLM, requires only two epochs to achieve convergence on the validation set, demonstrating the model’s rapid convergence rate. This quick convergence suggests that LLM-PSS can be trained efficiently in industrial settings. Furthermore, the training and validation error curves maintain an ideal gap throughout the training process, reflecting good generalization ability. The model performs well on both the training and validation sets, showcasing its robust performance across different data scenarios.

Due to the introduction of the text modality into its inputs, LLM-PSS exhibits a broader range of interpretability compared to traditional DDSS. To further explain and analyze LLM-PSS, we visualize the attention weights between all tokens in the input prompt. It is important to note that GPT-2 utilizes the byte pair encoding (BPE) \cite{sennrich-etal-2016-neural} tokenization algorithm, which breaks down prompts into subword-level tokens, potentially resulting in incomplete words within tokens. To enhance the visual clarity of our analysis, tokens with the same word subparts are concatenated into single tokens for heatmap visualization. In each heatmap plot, hotter regions indicate higher attention values.

\begin{figure*}[!t]
\centering
\subfloat[]{\includegraphics[width=2.5in]{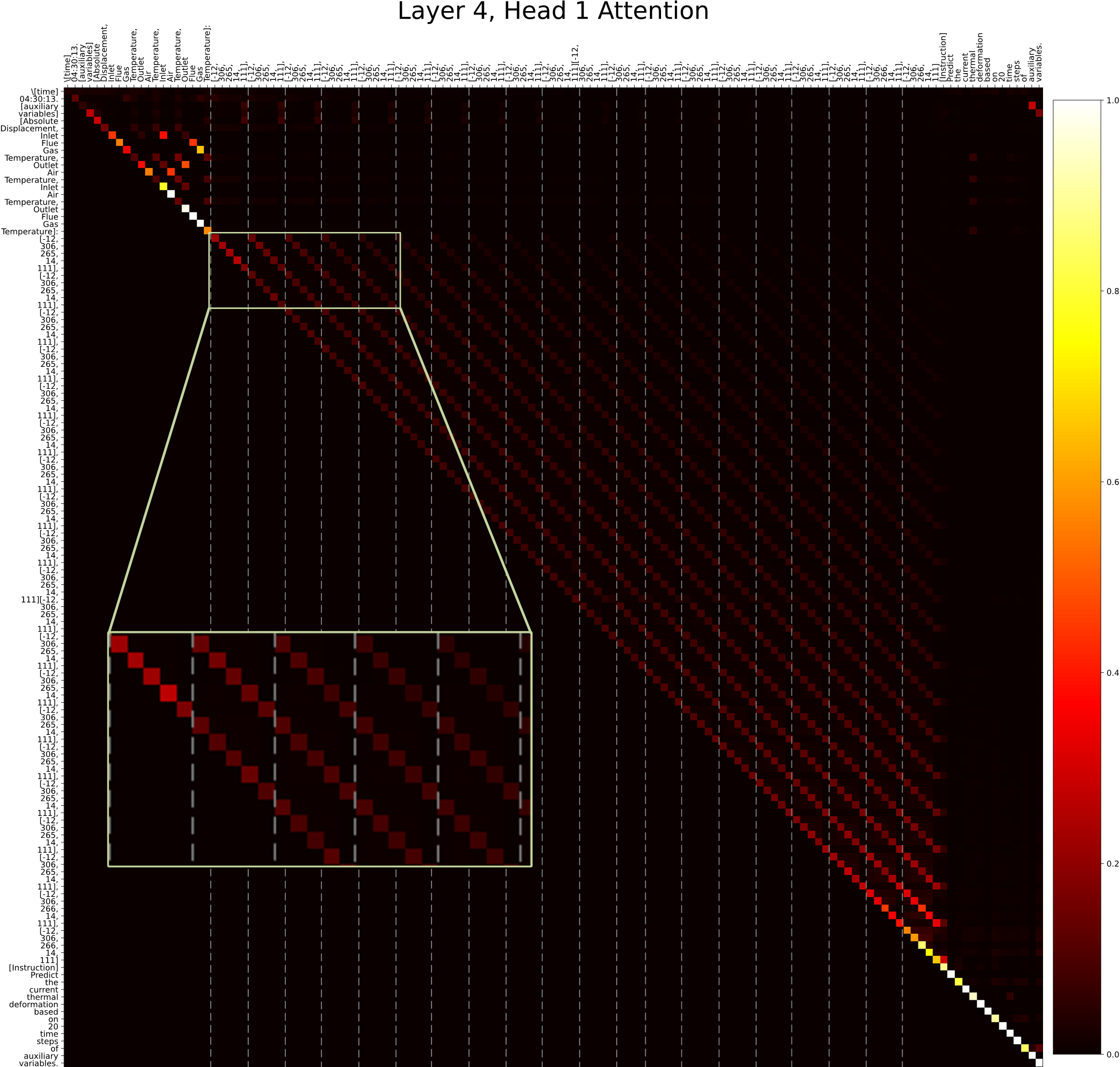}%
\label{fig19_first_case}}
\hfil
\subfloat[]{\includegraphics[width=2.5in]{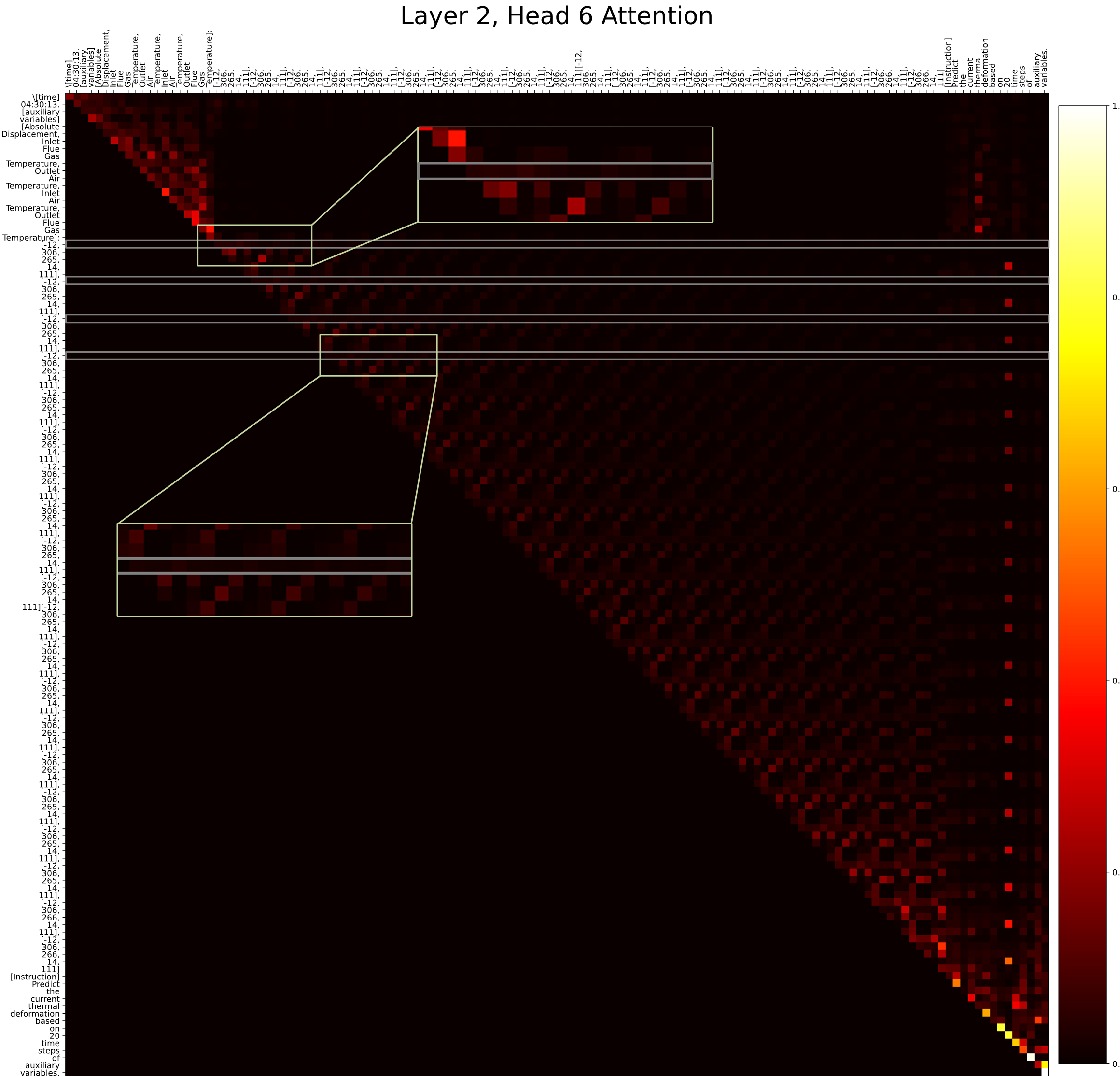}%
\label{fig20_second_case}}
\caption{The visualization results of attention heatmaps for LLM-PSS. (a) Case 1 (Layer 4, Head 1). (b) Case 2 (Layer 2, Head 6).}
\label{fig19}
\end{figure*}

For Case 1 (as shown in Fig. \ref{fig19}\subref{fig19_first_case}), it is evident that each auxiliary variable value at every time step pays attention to its own value at other time steps (the prompt text spans 20 time steps, delineated by gray dashed lines). This attention diminishes as the time interval increases, indicating that LLM-PSS can comprehend the temporal representation in the prompt text and effectively capture dependencies among values at different time steps. Additionally, the top-left section of the heatmap reveals the attention relationships among the values of AVs.

For Case 2 (as shown in Fig. \ref{fig19}\subref{fig20_second_case}), it is evident from the right side of the figure that the token "20 time steps" attends to the first auxiliary variable value at each time step, totaling 20 points of attention. This indicates that the introduced temporal information can be correlated with the AVs. Additionally, unlike in Case 1, where the focus was on the relationships within the same auxiliary variable across different time steps, Case 2 clearly highlights the attention relationships among different AVs. Furthermore, it can be observed that when the auxiliary variable value represents an absolute displacement (marked with gray boxes for the first three variables), its attention weights to other temperature-related auxiliary variable values are minimal, which aligns with the actual operating conditions of the air preheater \cite{9794453}.

\begin{figure}[!t]
\centering

\subfloat[]{\includegraphics[width=3.3in]{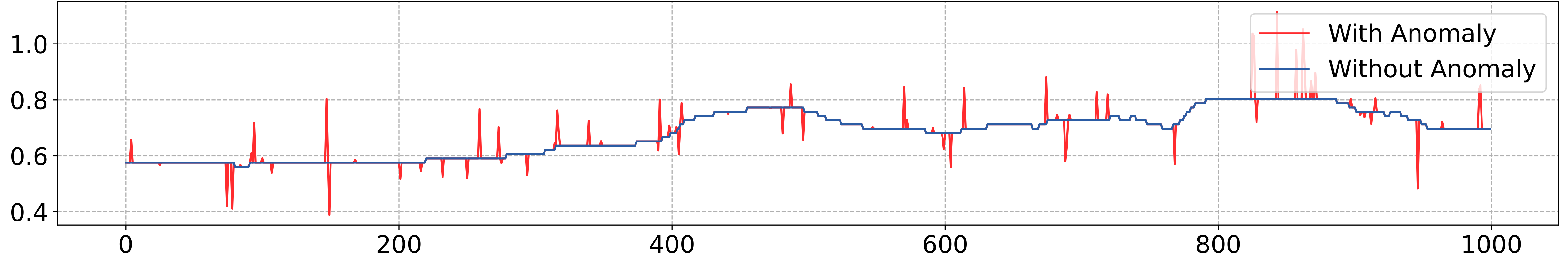}%
\label{fig21_first_case}}\\

\subfloat[]{\includegraphics[width=3.3in]{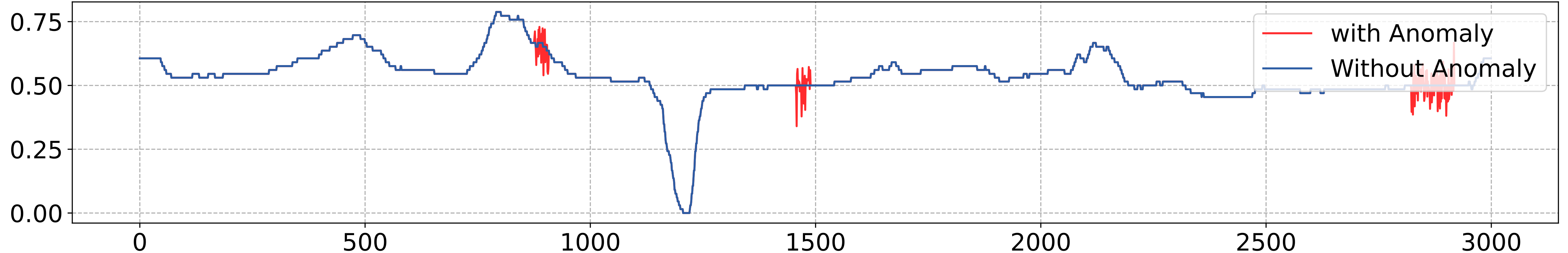}%
\label{fig22_second_case}}\\
\subfloat[]{\includegraphics[width=3.3in]{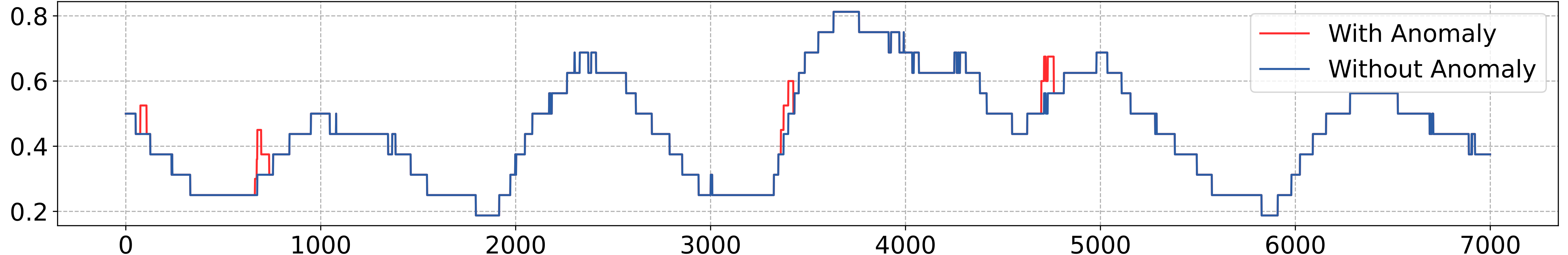}%
\label{fig23_third_case}}
\caption{Comparison of AVs series before and after anomaly injection. (a) Case 1: spike anomaly injected into Variable 1. (b) Case 2: noise anomaly injected into Variable 5. (c) Case 3: scale anomaly injected into Variable 4.}
\label{fig21}
\end{figure}

\subsection{Results of SSFM-TSA for Anomaly Detection}

We implemented the downstream anomaly detection task using SSFM-TSA through a semi-supervised sequence reconstruction method. To validate the effectiveness of SSFM-TSA in anomaly detection, we selected three typical anomalies common in industrial environments: spike anomaly, noise anomaly, and scale anomaly. First, as outlined in Section \ref{4.5}, we injected these three types of anomalies into each auxiliary variable, which includes the inlet temperature of the gas flue, the outlet temperature of the gas flue, the inlet temperature of the air duct, the outlet temperature of the air, and the absolute displacement, respectively represented as Variable 1 to Variable 5. Fig. \ref{fig21} illustrate the localized comparisons of these VAs before and after the injection of spike anomaly, noise anomaly, and scale anomaly.

\begin{table*}[t]
\centering
\caption{Full results for the anomaly detection.}
\label{table5}
\setlength{\tabcolsep}{3.5mm}{
\begin{tabular}{l|ccc|ccc|ccc|ccc}
\hline
\multirow{2}{*}{Variable} & \multicolumn{3}{c|}{Accuracy (\%)}&\multicolumn{3}{c|}{Recall (\%)}&\multicolumn{3}{c|}{Precision (\%)}&\multicolumn{3}{c}{F1-score (\%)} \\
\cline{2-13}

 & Noise & Spike & Scale  & Noise & Spike & Scale & Noise & Spike & Scale& Noise & Spike & Scale \\
\hline

%0.25x-----------------
Variable1    &98.53&	96.93	&97.27&	92.00	&60.45&	62.03&	75.11&	74.93	&45.01&	0.82&	0.67&	0.52\\

Variable2  	&98.77&	96.73	&97.45&	93.76&	69.57&	51.88&	70.90	&64.66&	51.27	&0.80&	0.67&	0.52\\

Variable3  		&99.38&	96.82	&98.15&	97.89	&71.26&	65.68	&85.07&	63.55	&54.23&	0.91&	0.67&	0.59	\\

Variable4		&99.36&	96.75	&98.21&	90.52&	70.98&	56.73&	77.84&	64.75	&46.79&	0.83&	0.68&	0.51\\

Variable5 &99.72&	98.42&	98.44&	97.86	&77.86&	56.85&	88.11&	73.99&57.43&	0.92&	0.76	&0.57\\
Average	&99.15&	97.13	&97.90&	94.41&	70.02&	58.63	&79.41&	68.38&	50.95&	0.86&	0.69	&0.54\\
\hline
\end{tabular}}
\label{table_MAP}
\end{table*}

We selected accuracy, precision, recall, and F1-score as metrics to validate the anomaly detection performance across all the aforementioned anomaly types. The results on the test set are presented in Table \ref{table5}. From the table, it can be observed that SSFM-TSA effectively detects most of the anomalies in various scenarios even though the original data was unlabeled.

\begin{table}[t]
\caption{\textbf{Missing value imputation results for different mask ratios.}}%标题
\label{table6}
\centering%把表居中

\setlength{\tabcolsep}{3.5mm}{
\begin{tabular}{lccc}
\toprule%第一道横线
Mask Ratio (\%)& MAE $\downarrow$ &  RMSE $\downarrow$ &MAPE (100\%) $\downarrow$\\
\midrule%第二道横线 
10\%&	0.0068&	0.0138&	1.4207\\
20\%	&0.0080&	0.0155&	1.6811\\
30\%	&0.0091&	0.0172&	1.8916\\
40\%&	0.0104&	0.0195&	2.1678\\
Average&	0.0086&	0.0165&	1.7903\\

\bottomrule%第三道横线
\end{tabular}}
\end{table}

\subsection{Results of SSFM-TSA for Imputation}

In the downstream task of missing value imputation within SSFM-TSA, we randomly masked 10\%, 20\%, 30\%, and 40\% of the time points for all AVs series to validate the robustness of the model. The experimental results are presented in Table \ref{table6}. The results indicate that our method consistently achieves satisfactory performance in predicting missing values across the four masking ratios. MAPE is used to quantify the error between the predicted values and the actual values, expressed as a percentage. Even with 40\% of the data missing, SSFM-TSA demonstrates a minimal error between the predicted values and the true values.

\begin{figure}[!t]
\centering
\subfloat[]{\includegraphics[width=1.7in]{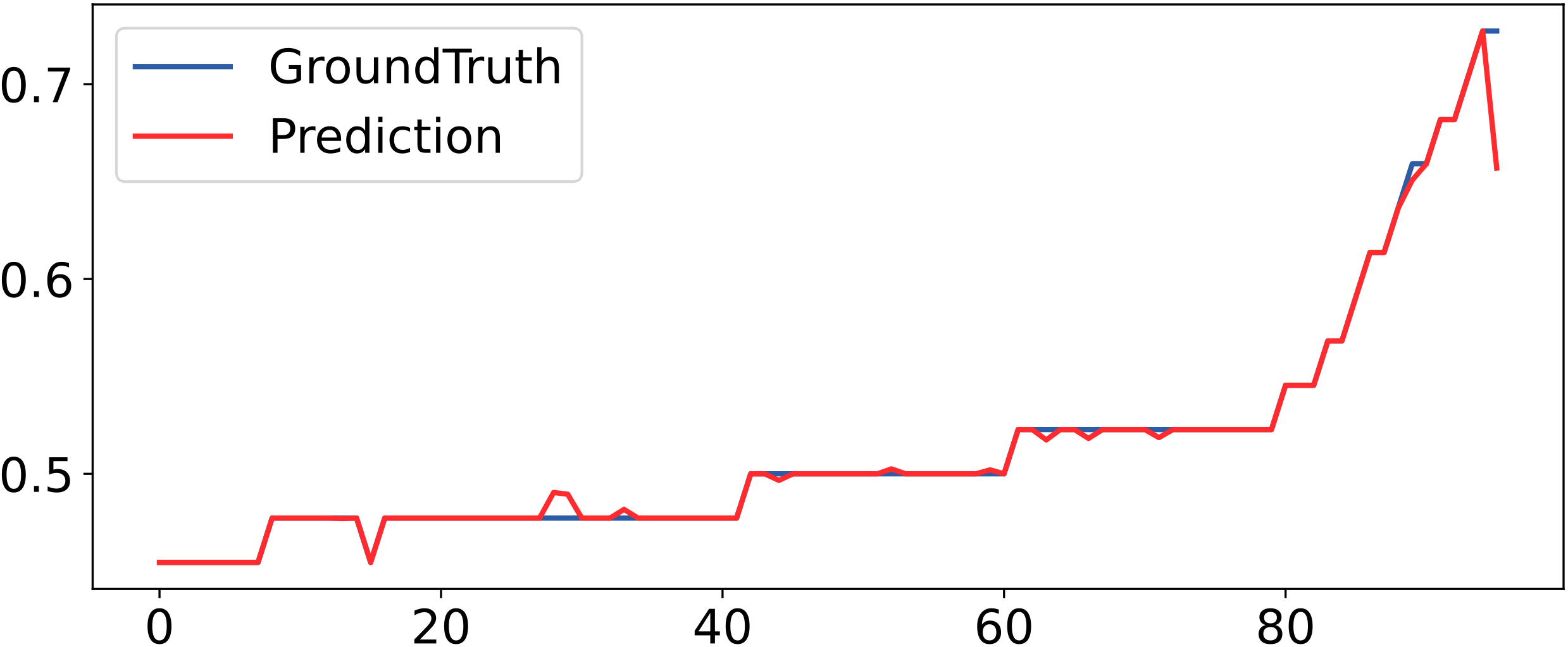}%
\label{fig_24_first_case}}
\hfil
\subfloat[]{\includegraphics[width=1.7in]{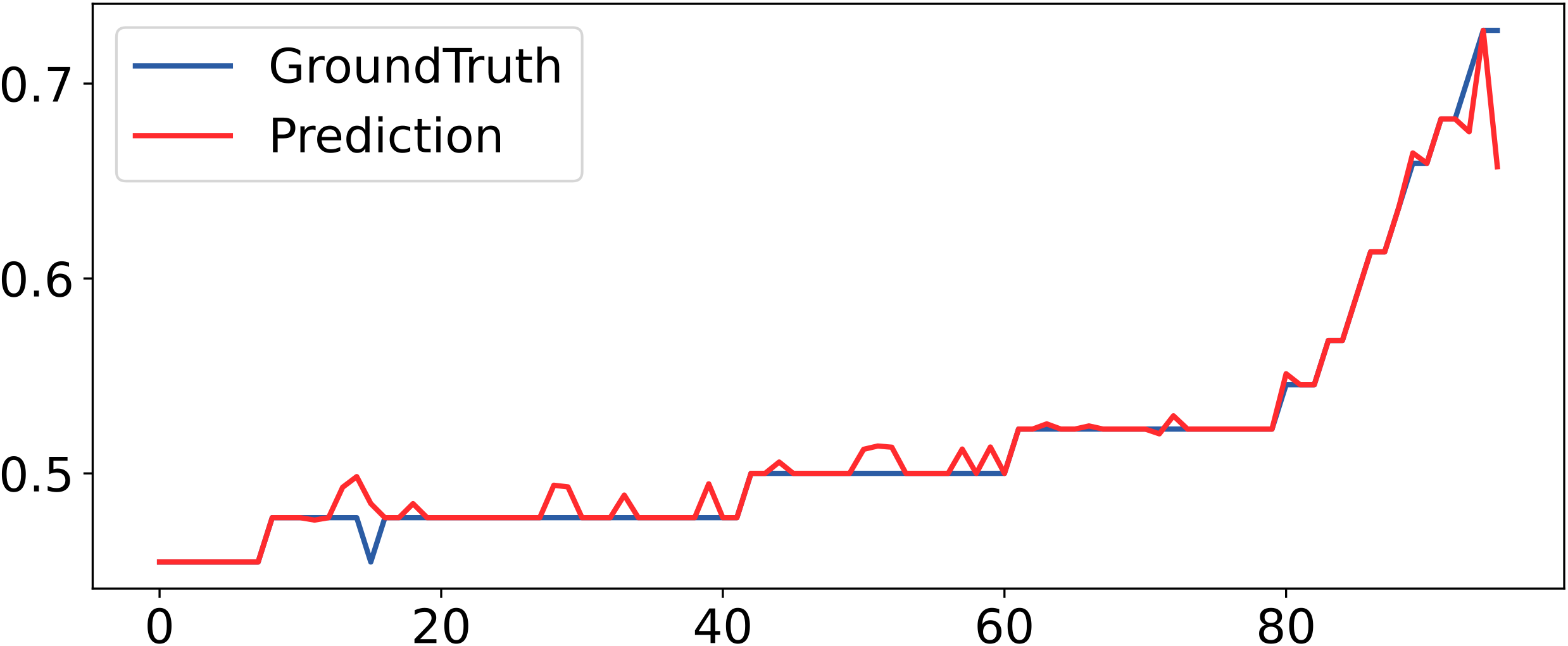}%
\label{fig_25_second_case}}

\subfloat[]{\includegraphics[width=1.7in]{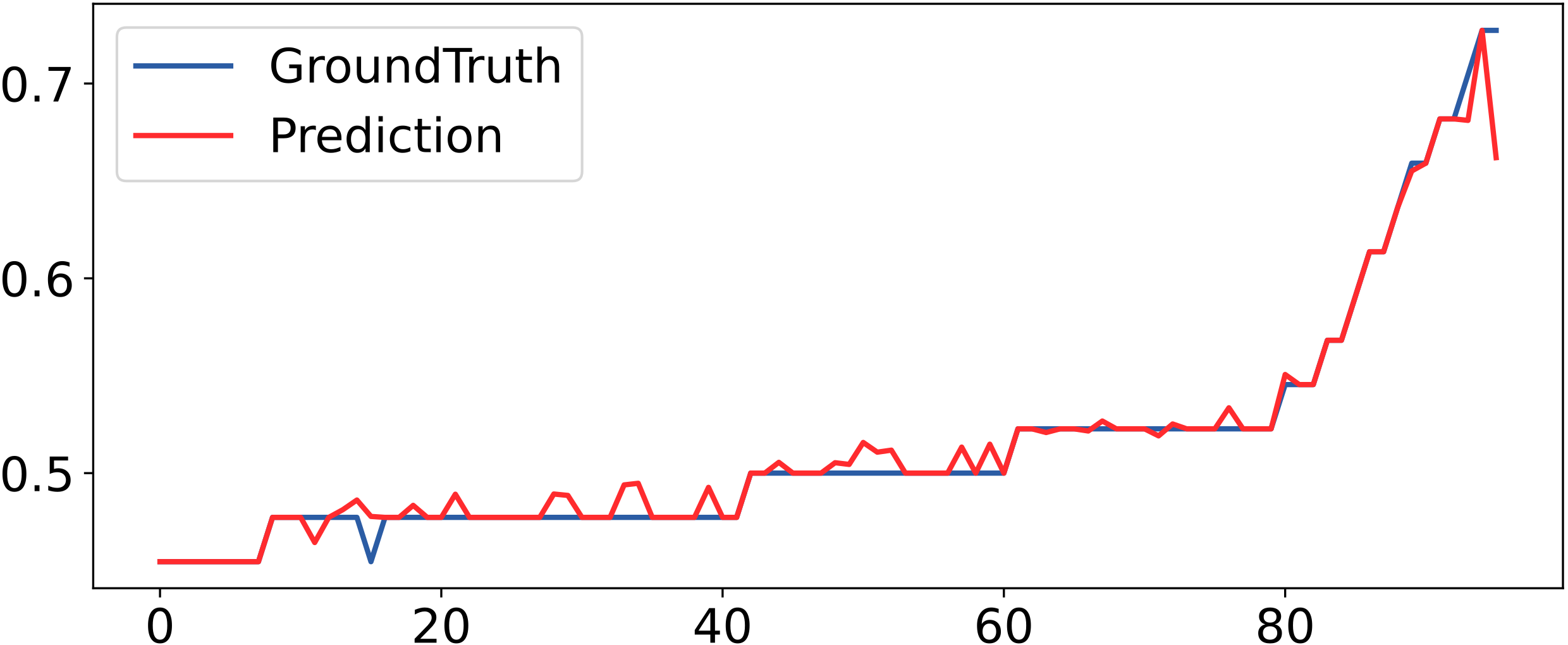}%
\label{fig_26_third_case}}
\hfil
\subfloat[]{\includegraphics[width=1.7in]{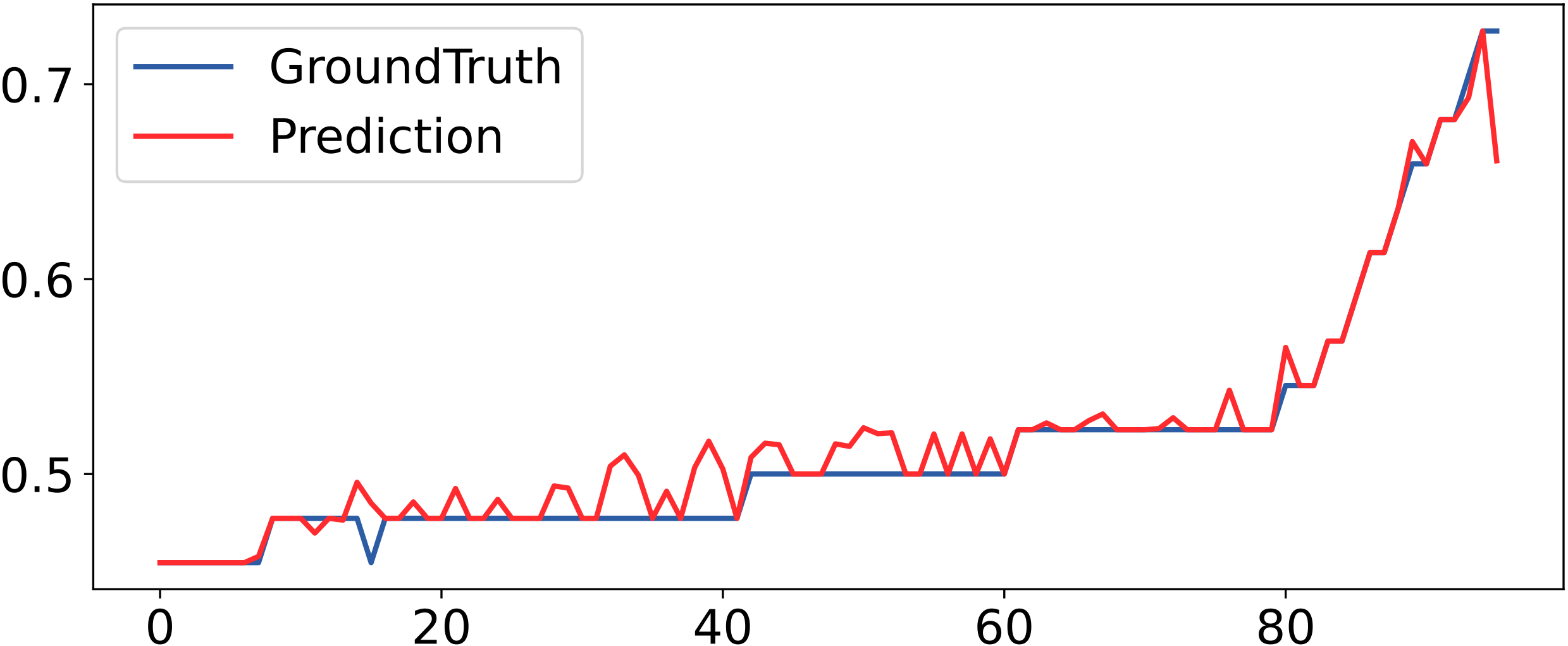}%
\label{fig_27_fourth_case}}

\caption{Missing value imputation results for a sample in Variable 5 under different mask ratios. (a) Mask ratio of 10\%. (b) Mask ratio of 20\%. (c) Mask ratio of 30\%. (d) Mask ratio of 40\%.}

\label{fig24}
\end{figure}

To visually demonstrate the predictive performance of SSFM-TSA under various mask ratios, as shown in Fig. \ref{fig24}, we selected a sample (96-time steps) from Variable 5 in the test set for qualitative analysis. The vertical axis is scaled between 0.45 and 0.70. The blue line represents the true series of the sample, while the red line depicts the predicted series after imputation with SSFM-TSA on masked data. From the graph, it is evident that the proposed method effectively tracks the true curve across mask ratios of 10\% to 40\%, maintaining prediction errors within a small range.

\subsection{Ablation Study}
To validate the effectiveness and contributions of key components in LLM-KETSS, as well as the enhancement of soft sensor performance by pretrained LLM, we conducted comprehensive ablation experiments in downstream tasks of DDSS. We designed five different variants for evaluation. Firstly, we introduced Variant 1 (W/O time series), where we reduced the original 96-time window length to 1, thereby reducing the number of tokens encoded by the AVS encoder to 1. Variant 2 (W/O Pre-training) omitted the strategy of using pretrained LLM followed by two-stage fine-tuning; instead, it involved full parameter training directly on the training dataset. Variant 3 (W/O Adapter) did not employ TSA in the second stage for adaptation training with downstream tasks; instead, it utilized the PEFT strategy to fine-tune LN layers and linear output layers. In Variant 4 (W/O Stage1 PEFT), we skipped the first-stage modal alignment training and directly adapted the pretrained LLM to downstream tasks in the second stage. Lastly, Variant 5 (WO AVS Encoder) did not utilize the AVS Encoder to process input sequences; instead, it flattened the input sequences into one-dimensional sequences, where each value served as a token, and then used linear layers to embed these input tokens into specific dimensions.

\begin{table}[t]
\caption{\textbf{Comparison of few-shot soft sensing prediction results on 10\% training data}}%标题
\label{table7}
\centering%把表居中

\begin{tabular}{lcccc}
\toprule%第一道横线
Model&SMAPE $\downarrow$& MAE $\downarrow$ &  RMSE $\downarrow$& ${{R}^{2}}$ $\uparrow$ \\
\midrule%第二道横线 
W/O Stage1 PEFT&6.4816&	0.0361&	0.0491&	0.8979\\
W/O Adapter&6.5745&	0.0369&	0.0503&		0.8926\\
W/O Pre-training&6.7397&	0.0366&	0.0494&		0.8972\\
W/O Time series&7.2533&	0.0393&	0.0513&		0.8883\\
W/O AVS Encoder&7.6290&	0.0401&	0.0513&		0.8883\\
\textbf{LLM-DSS (ALL)}&\textbf{5.9471}&	\textbf{0.0335}&	\textbf{0.0470}&	\textbf{0.9063}\\

\bottomrule%第三道横线
\end{tabular}

\end{table}

The results of the ablation experiments are presented in Table \ref{table7}, demonstrating that removing any component leads to a decline in overall performance of the proposed model. The experimental comparison in Variant 1 highlights the significant role of LLM in capturing long-range dependencies among tokens in industrial time series, crucial for soft sensing tasks. Variant 2 comparisons indicate that innate world knowledge embedded in pretrained LLM provides additional benefits for soft sensor. Results from Variant 3 suggest that using TSA is more effective in adapting to downstream tasks compared to PEFT. Variant 4 results validate the critical role of the first-stage modal alignment in subsequent adaptation tasks. Variant 5 results indicate that the proposed AVS Encoder effectively tokens and embeds AVs data.

\section{Conclusion}

In this article, to address the issues of poor general performance of DDSS, single-modal inputs, and weak capabilities with limited samples, we propose a universal soft sensor modeling framework leveraging the emergent capabilities of pretrained LLM. Initially, to bridge the domain gap between LLMs trained on natural language and sensor data, an AVS Encoder is introduced to encode auxiliary variable data into tokenized forms suitable for LLM inputs. Subsequently, through a two-stage fine-tuning and adaptation strategy, the SSFM achieves adaptive performance across various complex downstream soft sensor modeling tasks using the same model framework. This is achieved by fine-tuning a small set of parameters in the adapter, enabling rapid adaptation to high-performance downstream tasks. Furthermore, this article introduces LLM-PDSS and LLM-PSS, which encode text knowledge information based on prompts and structured data into a unified token format for knowledge embedding. Leveraging the transformer architecture of LLMs for multimodal input data training, experimental validation demonstrates that introducing text modality effectively assists soft sensor models in achieving more accurate predictions. Finally, our method is compared with others in a scenario with only 10\% of data samples, demonstrating robust predictive performance in small sample settings.

Future research will focus on leveraging the multimodal capabilities of LLMs to achieve more accurate and robust soft sensing modeling. Additionally, enhancing the interpretability of soft sensing through the introduction of natural language modality to meet the safety requirements for deployment in process control systems is an area worthy of further study.

\bibliographystyle{ieeetr}
\bibliography{IEEEabrv,ref}
\end{document}